\documentclass[11pt]{article}

\usepackage[preprint]{acl}

\usepackage{times}
\usepackage{latexsym}
\usepackage[T1]{fontenc}
\usepackage[utf8]{inputenc}
\usepackage{microtype}
\usepackage{inconsolata}
\usepackage{graphicx}
\usepackage{booktabs}
\usepackage{array}
\usepackage{amsmath}
\usepackage{verbatim}
\usepackage{natbib}
\usepackage{colortbl}
\usepackage{xcolor}
\usepackage{longtable}
\usepackage{amssymb}
\newcolumntype{L}[1]{>{\raggedright\arraybackslash}p{#1}}

\usepackage[textsize=scriptsize,backgroundcolor=green,disable]{todonotes}
\definecolor{emptyrow}{gray}{0.92}

\title{From Interpretability to Control: Insights from Six Years of the TrustNLP Workshop}

\author{
  \textbf{Rahul Gupta\textsuperscript{1}, \,
  Abhinav Mohanty\textsuperscript{1}, \,
  Anaelia Ovalle\textsuperscript{2}, \,
  Anil Ramakrishna\textsuperscript{2}} \\
  \textbf{Anubrata Das\textsuperscript{3}, \,
  Apurv Verma\textsuperscript{4}, \,
  Jwala Dhamala\textsuperscript{1}, \,
  Ninareh Mehrabi\textsuperscript{2}} \\
  \textbf{Tharindu Kumarage\textsuperscript{1}, \,
  Yada Pruksachatkun\textsuperscript{5}, \,
  Yang Trista Cao\textsuperscript{1}} \\
  \textbf{Kai-Wei Chang\textsuperscript{6}, \,
  Aram Galstyan\textsuperscript{1}} \\[0.6em]
  \normalfont
  \textsuperscript{1}Amazon AGI \quad
  \textsuperscript{2}Meta \quad
  \textsuperscript{3}Autodesk \quad
  \textsuperscript{4}New Jersey Institute of Technology \\
  \textsuperscript{5}Salesforce \quad
  \textsuperscript{6}University of California, Los Angeles \\
}

\begin{document}
\maketitle
\begin{abstract}
The Workshop on Trustworthy Natural Language Processing (TrustNLP), co-located with major ACL conferences since 2021, has grown from 8 proceedings papers to 41 over six editions, documenting a field-wide transition from post-hoc interpretability of static models to mechanistic understanding and proactive control of generative systems. We synthesize insights from all 144 proceedings papers, classifying them along six trust dimensions grounded in established frameworks (TrustLLM, DecodingTrust). We observe co-occurrences with capability emergence. The release of the first high-impact chat models activated all trust dimensions simultaneously, while subsequent model generations shifted focus toward truthfulness and safety alignment. Analysis from the classification study reveals that truthfulness is the fastest-growing dimension (absent in 2021--2022, comprising 37\% of papers by 2025--2026), fairness remains the most consistent theme, and explainability exhibits a U-shaped trajectory; declining as post-hoc methods lost relevance but resurging in 2026 through mechanistic interpretability. A cross-venue comparison with ACL, NAACL, EACL, and EMNLP ($\sim$ 2K papers) in the same period shows that TrustNLP's topical distribution closely follows the field average. We identify four structural insights and conclude with actionable directions for the research community.
\end{abstract}

\section{Introduction}

\begin{figure}[t]
\centering
\includegraphics[width=\columnwidth]{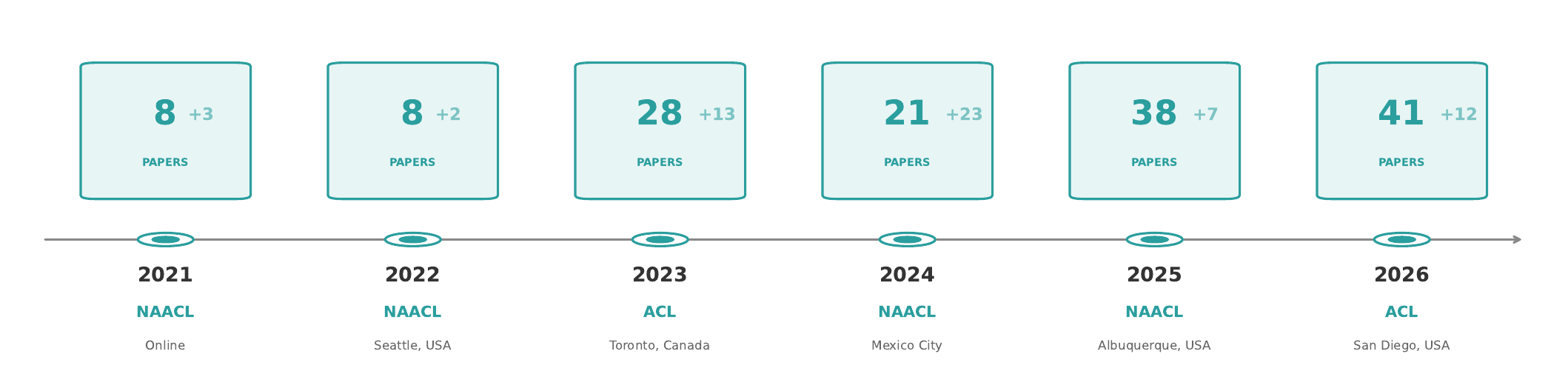}
\caption{TrustNLP workshop series timeline (2021--2026). Numbers in lighter shade indicate additional non-archival accepted papers; this paper analyzes only the archival proceedings. All editions use double-blind peer review with archival and non-archival tracks.}
\label{fig:timeline}
\end{figure}

The rapid deployment of LLMs in critical infrastructure has made trustworthiness a first-order research concern. The TrustNLP workshop series, co-located with NAACL and ACL since 2021, has served as a key ACL-community venue for trustworthy NLP research \cite{trustnlp-2021-trustworthy,trustnlp-2022-trustworthy,trustnlp-2023-trustworthy,trustnlp-2024-trustworthy,trustnlp-2025-trustworthy}. The workshop is organized by a stable core of academic and industry institutions. Figure~\ref{fig:timeline} presents a timeline of the growth of the workshop. Alongside the 5$\times$ increase in archived proceedings papers between 2021 and 2026, the research focus underwent a qualitative transformation in what ``trust'' means. 
 Before the era of LLM launches ($\sim$ November 2022), the key questions were around \emph{fairness and interpretability}: can we audit model decisions? After the launch of high-impact chat models, trust also became about \emph{reliability and trade-offs}: is the LLM consistent across prompts, and how do fairness, performance, and explainability interact? By 2025--2026, with multimodal \& agentic systems, trust evolved into \emph{controllability}: can we constrain model behavior under adversarial conditions and mechanistically verify alignment?
%Before the era of LLM launches ($\sim$ November 2022), the key questions were around \emph{fairness and interpretability}: can we audit model decisions? After such a launch, trust also became about \emph{reliability}: is the LLM behavior consistent across prompt variations? By 2024--2025, with multimodal \& agentic systems, trust evolved into \emph{controllability}: can we constrain model behavior under adversarial conditions?

%This paper synthesizes insights from all six editions, arguing that trustworthy NLP evolves in \emph{punctuated equilibria} driven by capability emergence rather than steady progress. 
We make three contributions: (1)~a quantitative topic analysis of all 144 proceedings papers revealing the fastest-growing and most persistent research themes; (2)~a chronological synthesis of technical contributions organized by the dominant paradigm of each phase; and (3)~the identification of four structural insights and actionable directions for the research community.

We focus on TrustNLP as a representative case study as three properties make TrustNLP particularly well-suited for this role. First, its publication pipeline is integrated with the ACL ecosystem: accepted papers are peer-reviewed through ARR or direct submission and appear in the ACL Anthology alongside main-conference proceedings at NAACL, ACL, and EACL, ensuring that the work reflects the quality norms and topical pulse of the broader NLP community. Second, the workshop maintains a deliberately narrow scope centered on trustworthiness challenges in generative and interactive language systems, rather than serving as a venue for AI ethics broadly construed; this focus yields a sharper signal about how the field responds to specific capability emergence. Third, TrustNLP offers longitudinal continuity with six consecutive editions (2021–2026) with a stable core of organizers and a consistent programmatic agenda. This enables us to trace thematic evolution over time.

\begin{comment}
\begin{table}[t]
\centering
\small
\begin{tabular}{@{}llrl@{}}
\toprule
\textbf{Year} & \textbf{Venue} & \textbf{Papers} & \textbf{Dominant Paradigm} \\
\midrule
2021 & NAACL (Online) & 9 & Interpretability \\
2022 & NAACL (Seattle) & 9 & Fairness auditing \\
2023 & ACL (Toronto) & 29 & LLM reliability \\
2024 & NAACL (Mexico City) & 22 & Adversarial control \\
2025 & NAACL (Albuquerque) & 39 & Multimodal safety \\
\bottomrule
\end{tabular}
\caption{TrustNLP workshop editions and dominant research paradigms. The 2023 edition received 57 submissions (41 accepted, 29 in proceedings); 2024 received 44 submissions (40 accepted, 22 in proceedings).}
\label{tab:overview}
\end{table}
\end{comment}

\section{Related Work}

The study of trustworthiness in AI systems spans multiple communities, each contributing distinct methodological perspectives. We situate TrustNLP within this broader landscape by surveying related workshops, benchmark efforts, and analysis articles that collectively define the field.

Several comprehensive surveys and benchmarks have sought to systematize the evaluation of trustworthiness in large language models. DecodingTrust~\citep{wang-etal-2024-decodingtrust} provides a multi-dimensional assessment of GPT models across toxicity, stereotype bias, adversarial robustness, out-of-distribution generalization, privacy, machine ethics, and fairness, establishing one of the first holistic evaluation protocols for proprietary models. TrustLLM~\citep{huang2024trustllm} proposes a unified evaluation framework that encompasses truthfulness, safety, fairness, robustness, privacy, machine ethics, transparency, and accountability, and benchmarks over a dozen open-source and proprietary models to reveal systematic gaps between the stated alignment goals and measured behavior. \citet{liu-etal-2024-trustworthy-llms-survey} surveys the broader landscape from principles to practices, offering a taxonomy that connects technical methods, such as differential privacy, certified robustness, and calibration, to real-world deployment considerations including regulatory compliance and organizational governance. 
In addition, various works have surveyed the risks of LLM deployment in domains such as medicine~\citep{liu2024survey}, finance~\citep{chen2025standard}, education~\citep{harvey2025don} and cybersecurity~\citep{ahi2025large}. 
While these frameworks provide valuable snapshots of model trustworthiness at a single point in time, TrustNLP's longitudinal proceedings capture how trust challenges emerge, intensify, and transform in response to capability shifts, providing a complementary, diachronic perspective that static benchmarks cannot offer.

\section{Quantitative Topic Analysis}

To study how research topics have evolved across six editions, we classify all 144 archived proceedings papers by \emph{trust dimension}. That is, what aspect of trustworthiness does the paper primarily address?

\subsection{Classification Taxonomy}

We derive our trust taxonomy from TrustLLM~\citep{huang2024trustllm}, which benchmarks six dimensions of LLM trustworthiness: Truthfulness, Safety, Fairness, Robustness, Privacy, and Machine Ethics. We adopt these as our starting point and refine them using DecodingTrust's~\citep{wang-etal-2024-decodingtrust} finer-grained distinctions to clarify scope boundaries. DecodingTrust's \emph{Toxicity} is folded into Fairness \& Bias to merge commonalities where toxic content may disproportionately target particular marginalized groups. DecodingTrust's three robustness sub-dimensions inform the scope of our Robustness category. We argue that TrustLLM's \emph{Safety} and \emph{Machine Ethics} are methodologically intertwined and merge them into a single Machine Ethics \& Safety dimension. In this way, we expand the scope of each class by merging overlapping class labels from the two taxonomies.

Finally, neither TrustLLM nor DecodingTrust benchmarks explainability (TrustLLM proposes ``Transparency'' as a principle but does not operationalize it). However, explainability dominates the TrustNLP corpus in 2021--2022 and appears explicitly in the workshop CfP, so we add it as a corpus-motivated dimension.

This yields six trust dimensions: \emph{Fairness \& Bias}, \emph{Robustness \& Adversarial}, \emph{Privacy}, \emph{Machine Ethics \& Safety}, \emph{Truthfulness}, and \emph{Explainability}. Papers may belong to multiple dimensions.

\subsection{Annotation Methodology}

We used three independent annotation sources to classify all 144 papers. First, we defined annotation instructions specifying the scope of each trust dimension (reproduced in Appendix~\ref{sec:appendix-annotation-instructions}). Labels were then obtained from three sources: (i)~one author annotated all papers based on their title and abstract following the instructions; (ii)~Claude Sonnet~5 (Anthropic) and Amazon Nova Lite~2.0 each independently annotated the same papers using the same instructions as a system prompt (see Appendix~\ref{sec:appendix-llm-validation} for the exact prompts). 
After the first round, 60\% of papers (87/144) had perfect agreement across all three annotators, and in 78\% of cases (112/144) at least one LLM fully agreed with the human (please note that a paper may belong to multiple of the six labels). 
As per Table~\ref{tab:agreement}, agreement is substantial to excellent ($\kappa > 0.7$) for four of six dimensions. The two most contested boundaries---Machine Ethics \& Safety and Truthfulness---reflect a conceptual overlap (e.g., a paper on LLM safety degradation under attacks touches both robustness and safety).

\begin{table}[t]
\centering
\small
\resizebox{\columnwidth}{!}{%
\begin{tabular}{@{}lcc@{}}
\toprule
\textbf{Trust Dimension} & \textbf{Sonnet 5 ($\kappa$)} & \textbf{Nova Lite 2 ($\kappa$)} \\
\midrule
Fairness \& Bias & 0.892 & 0.782 \\
Robustness \& Adversarial & 0.749 & 0.706 \\
Privacy & 0.881 & 0.882 \\
Machine Ethics \& Safety & 0.532 & 0.506 \\
Truthfulness & 0.645 & 0.673 \\
Explainability & 0.737 & 0.647 \\
\midrule
\emph{Overall accuracy} & \emph{92.2\%} & \emph{90.5\%} \\
\bottomrule
\end{tabular}%
}
\caption{Cohen's $\kappa$ between human annotations and two LLM classifiers on trust dimensions (prior to adjudication). Agreement ranges from moderate (Machine Ethics \& Safety) to excellent (Privacy, Fairness \& Bias), with an overall decision-level accuracy above 90\%.}
\label{tab:agreement}
\end{table}

The human annotator then reviewed the remaining 40\% of cases with imperfect agreement and decided the final label, yielding the labels reported in Table~\ref{tab:topic-analysis}.
The final set of annotations (post second human review and those from models) are presented in Appendix~\ref{sec:appendix-annotation-review}.
Appendix~\ref{sec:appendix-crossvenue} presents an analysis on Trust papers across other *CL venues. A similar classification on trust papers (first filtered via keyword search), shows that TrustNLP's topical distribution generally matches with the broader field. This is indicative that the workshop mirrors the community's trust priorities,

\begin{table}[t]
\centering
\small
\resizebox{\columnwidth}{!}{%
\begin{tabular}{@{}lccccccc@{}}
\toprule
\textbf{Dimension} & \textbf{'21} & \textbf{'22} & \textbf{'23} & \textbf{'24} & \textbf{'25} & \textbf{'26} & \textbf{Total} \\
\midrule
Fairness \& Bias & 2 & 3 & 6 & 6 & 7 & 6 & 30 \\
\addlinespace
Robustness \& Adversarial & 0 & 0 & 6 & 5 & 8 & 8 & 27 \\
\addlinespace
Privacy & 1 & 0 & 6 & 0 & 1 & 1 & 9 \\
\addlinespace
Machine Ethics \& Safety & 0 & 2 & 3 & 3 & 3 & 9 & 20 \\
\addlinespace
Truthfulness & 0 & 0 & 6 & 5 & 14 & 13 & 38 \\
\addlinespace
Explainability & 5 & 4 & 2 & 3 & 2 & 13 & 29 \\
\bottomrule
\end{tabular}%
}
\caption{Trust dimension distribution across six TrustNLP editions. Dimensions are not mutually exclusive; papers spanning multiple areas are counted in each relevant category. Dimensions are derived from TrustLLM, enriched with DecodingTrust scope boundaries, plus Explainability as a corpus-motivated addition.}
\label{tab:topic-analysis}
\end{table}

\subsection{Trends}

Several patterns emerge from the trust dimension distribution:

\textbf{Truthfulness} is the fastest-growing dimension, absent entirely in 2021--2022 but rising to 13--14 papers per year in 2025--2026 (38 total, 26\% of the corpus). This surge tracks the LLM era: as generative models became the dominant paradigm, hallucination, factuality, and calibration emerged as first-order trust concerns~\cite{shi-etal-2025-ambiguity, rawte-etal-2025-factoid}.

\textbf{Fairness \& Bias} is the most stable dimension, present in every edition at 2--7 papers/year (30 total). However, its nature has shifted: early work focused on standalone bias measurement~\cite{matthews-etal-2021-gender, azarpanah-farhadloo-2021-measuring}, while recent work reveals fairness--explainability trade-offs~\cite{brandl-etal-2024-interplay}, fairness--performance trade-offs~\cite{bui-von-der-wense-2024-trade}, and internal-vs-output bias mismatches. Fairness is transitioning from a standalone objective to a constraint in multi-objective optimization.

\textbf{Robustness \& Adversarial} was absent in 2021--2022 but stabilized at 5--8 papers/year from 2023 onward (27 total), driven by jailbreaks~\cite{kim-cho-2025-break}, red-teaming~\cite{lal-etal-2024-automated}, and multimodal attacks~\cite{cheng-etal-2025-pbi}. The nature of adversarial work has shifted: pre-LLM work focused on perturbing inputs, while post-LLM work involves co-constructing attacks \emph{with} the model itself~\cite{fu-etal-2024-cross}.

\textbf{Machine Ethics \& Safety} shows steady growth from 0 papers in 2021 to 9 in 2026 (20 total), reflecting the post large LLM (such as ChatGPT) focus around alignment, refusal behavior, and safety guardrails.

\textbf{Explainability} peaked early (9 papers across 2021--2022), declined to 2 papers in 2025, but rebounded sharply to 13 in 2026 (29 total). The early peak reflects the pre-LLM focus on understanding model decisions; the 2026 resurgence is driven by mechanistic interpretability---sparse autoencoders, linear probes, and activation steering---representing a shift from post-hoc explanations to causal understanding of model internals.

\textbf{Privacy} appears sporadically (concentrated in 2023 with 6 papers; \citealp{arnold-etal-2023-driving}), suggesting that privacy remains a specialized concern within the TrustNLP community despite its prominence in the broader trustworthiness literature.

%A cross-venue comparison with ACL, NAACL, EACL, and EMNLP (Appendix~\ref{sec:appendix-crossvenue}) shows that a similar distributional pattern holds across the broader *CL community, with TrustNLP's topical mix converging to near-identity with the field average by 2024--2026.

\section{Evolution of Technical Contributions}

While the previous section analyzed how individual topics evolved over time, this section examines how the research emphasis of each TrustNLP edition shifted year over year.

\subsection{Phase 1: Interpretability and Bias (2021--2022)}

The inaugural workshops focused on fairness, explainability, and privacy \cite{trustnlp-2021-trustworthy,trustnlp-2022-trustworthy}. As large pre-trained models became the predominant paradigm around this time, we observed a shift toward research focused on measuring representational harms in generative models. During this period, NLP tasks were predominantly classification and ranking systems \cite{devlin-etal-2019-bert, wang-etal-2018-glue}, and trust was fundamentally about transparency of decisions. Two complementary research threads defined this phase: establishing interpretability as an auditing requirement beyond simple accuracy reporting, and exposing how demographic and cross-lingual biases are structurally embedded in popular corpora and model representations.

\paragraph{Interpretability and Error Characterization}
Research established interpretability as a test for auditing decisions in consequential domains such as healthcare and legal systems. Early work explored jointly bootstrapping neural relation extractors with explanation decoders to ensure that outputs were accompanied by human-readable rationales \cite{tang-surdeanu-2021-interpretability}. The challenge of \emph{accountable error characterization} was central: researchers argued that simply reporting accuracy is insufficient for trust; models must quantify their own uncertainty. This led to attention-based attribution frameworks that identify features responsible for both performance and fairness through intervention and weight manipulation \cite{balkir-etal-2022-challenges}.

These attribution frameworks provided a mechanism for auditing not just \emph{what} a model predicts, but \emph{why}, a distinction that would become increasingly important as models grew more opaque. The emphasis on uncertainty quantification in these early papers anticipated the later focus on truthfulness \& confidence calibration that would dominate the 2025 edition. Keynote addresses during this period reinforced and broadened these themes: Wang discussed the evolution of ML in clinical data, cautioning against ``Frankenstein datasets,'' datasets assembled from other datasets and redistributed under new names, which can introduce duplication, source overlap, and biased, non-generalizable performance estimates~\cite{roberts2021common}, while Yang examined the social dimensions of trustworthy language technologies. She presented work on users' perceived trust in AI-mediated writing across social contexts and positive reframing techniques that neutralize negative viewpoints while preserving original meaning~\cite{ziems-etal-2022-inducing,hovy-yang-2021-importance}.

In 2022, researchers probed the Allen AI Delphi model to investigate moral reasoning \cite{fraser-etal-2022-moral}. They found that models mirror the moral principles of the demographic groups involved in the annotation process, raising fundamental questions about whose values are codified in ``moral'' AI systems. The \emph{Encoder Marginalization} framework was also introduced for dense passage retrieval, quantifying the contribution of different encoders to in-domain accuracy and providing a mechanism for identifying affecting factors during training \cite{li-etal-2022-encoder}. This early work on encoder analysis foreshadowed the mechanistic interpretability research that would emerge in 2026.

\paragraph{Cross-Lingual and Demographic Bias} A notable contribution in 2021 was a cross-linguistic study of gender bias in Wikipedia corpora across nine languages, including Arabic, German, Farsi, and Urdu \cite{matthews-etal-2021-gender}. This work showed that gender bias in NLP corpora is not solely an English-language concern and that bias metrics developed for English require careful adaptation when applied across typologically diverse languages. Related work on word embeddings further demonstrated that measured bias can vary substantially depending on the choice of similarity measure and descriptive statistic, underscoring the need for careful methodological choices when evaluating demographic bias in embedding spaces \cite{azarpanah-farhadloo-2021-measuring}.

\subsection{Phase 2: The Generative Pivot (2023)}

The 2023 proceedings mark the workshop's first major expansion after the release of large LMs like ChatGPT, growing from 8 papers in 2022 to 28 in 2023 \cite{trustnlp-2023-trustworthy}. The capability emergence activated research across several trust dimensions simultaneously. While fairness remained steady (6 papers), robustness, privacy, and truthfulness all saw first-time emergence---each contributing 6 papers, making 2023 the most evenly diversified edition. We discuss contributions across truthfulness, robustness \& privacy, and fairness \& toxicity next.

\paragraph{Truthfulness}
With generative models deployed at scale, truthfulness emerged as a first-order concern. Work on abstractive summarization addressed the familiar problem that generated summaries may be unsupported by, or inconsistent with, the source document, placing factuality at the intersection of model training and evaluation \citep{chern-etal-2023-improving}. Other papers asked how reliability should be assessed when model behavior depends on prompt wording, when benchmark data may have appeared in training data, and when confidence must be inferred from patterns of agreement across prompts \citep{khatun-brown-2023-reliability,aiyappa-etal-2023-trust,portillo-wightman-etal-2023-strength}. Together, these papers show the workshop moving from general concerns about unreliable generation toward prompt-robust and contamination-aware evaluation.

\paragraph{Robustness \& Privacy}
Robustness went from zero papers in 2021--2022 to six in 2023: work on sample attackability and minority-language robustness treated trust as resistance to perturbation and attack, including for languages and scripts outside the usual English-centered evaluation setting \citep{raina-gales-2023-sample,cao-etal-2023-pay-attention}. Privacy similarly surged, with papers examining personal information in training corpora and pseudonymization as a way to preserve data utility while reducing exposure risk \citep{subramani-etal-2023-detecting,yermilov-etal-2023-privacy}. These papers broadened the generative-model discussion by emphasizing the data, associated privacy threat models, and deployment conditions around the system, not only the surface quality of generated text.

\paragraph{Fairness \& Toxicity}
The workshop's earlier fairness concerns extended into the generative era. Papers on abuse detection, disability bias, representational-harm metrics, and name-based effects examined how models can treat social groups unevenly and how such behavior can be measured or diagnosed \citep{yee-etal-2023-keyword,narayanan-venkit-etal-2023-automated,hosseini-etal-2023-empirical,jeoung-etal-2023-examining}. Fairness maintained its steady presence (6 papers), but the methods shifted from measuring pre-trained embeddings to auditing generative outputs. The overall lesson of 2023 is that capability emergence activated multiple trust concerns in parallel rather than concentrating attention on any single dimension.

\subsection{Phase 3: Trust as a Trade-off Problem (2024)}

With rapid improvements in model quality \cite{anthropic2023claude2, openai2023gpt4} and progress in open models \cite{touvron2023llama2}, the 2024 edition shifted from studying trust dimensions in isolation to understanding how they interact under deployment constraints \cite{trustnlp-2024-trustworthy}. The diverse focus from 2023 continued: Fairness (6 papers), Robustness (5), and Truthfulness (5) remained nearly balanced, though Privacy declined from its 2023 peak to zero papers. We call out three themes from the proceedings: machine-generated text detection, proactive safety, and the emergence of multi-dimensional trade-offs.

\paragraph{Machine-Generated Text Detection}
One study compared shallow learning, LM fine-tuning, and multilingual fine-tuning for detecting machine-generated text \citep{adilazuarda-2024-beyond}. Another showed that detection scores are not determined only by the generated text: prompts and real texts can affect them too \citep{yoo-etal-2024-exploring}. Using causal diagrams, the authors identified backdoor paths and showed that some confounding bias can be partly reduced. This suggests that future detectors should account for prompt and context effects, not only classify outputs in isolation.

\paragraph{Proactive Safety and Red-Teaming}

Safety work became more proactive. \emph{Automated Adversarial Discovery} framed red-teaming as a search for attacks that both fool a safety classifier and belong to previously unseen harm dimensions \citep{lal-etal-2024-automated}. In parallel, \emph{Cross-Task Defense} showed that instruction tuning with refusal examples can help LLMs handle malicious long documents, such as manuals for illicit activities, while still performing benign NLP tasks \citep{fu-etal-2024-cross}. 
%The shared lesson is simple: safety systems must be trained not only on known failures, but against failures that have not yet been seen.

\paragraph{Multi-Dimensional Trade-offs}

The defining contribution of 2024 was the recognition that trust dimensions conflict. Adapter modules kept text classification accuracy close to full fine-tuning while reducing training time, but their fairness effects varied by sensitive group and could amplify existing bias in some settings \citep{bui-von-der-wense-2024-trade}. Fairness did not move cleanly with explainability: methods that improved one dimension did not reliably improve the other \citep{brandl-etal-2024-interplay}. A lesson emerged wherein a model should not be called trustworthy because it improves on one axis alone; its deployment costs, group-level effects, and explanations need to be checked together. 
%This insight---that trustworthiness is a multi-objective optimization problem, not a scalar---would inform the field's subsequent development.

\subsection{Phase 4: Agentic \& Multimodal Frontiers (2025)}

The fifth workshop widened the trust problem beyond text-only generation \cite{trustnlp-2025-trustworthy}, asking what breaks when models are used with databases, unclear questions, images, audio, and evidence-heavy claims. We call out three themes: specialized deployment tests, multimodal privacy and attack surfaces, and low-cost checks that make answers or refusals easier to trust.

\paragraph{Specialized Deployment Tests} An IoT smart-building study tested whether models could turn questions into database queries \citep{pavlich-etal-2025-beyond}. Models improved at producing queries, but reasoning over returned rows (e.g., whether traffic looks malicious) remained hard, showing that query generation is only part of the job. For question answering, \citet{shi-etal-2025-ambiguity} detect ambiguity before answering by measuring disagreement across candidate answers rather than asking the model to label questions as ambiguous, which also tightens confidence calibration on the answerable cases.

\paragraph{Multimodal Privacy and Attacks} Audio privacy work on membership inference for CLAP models proposed a text-only detector that uses generated gibberish rather than exposing real audio to the target model \citep{cheng-etal-2025-gibberish}, shifting the question from whether a model repeats private data to whether we can tell a speaker was in training without leaking more about them. On the safety side, PBI-Attack \citep{cheng-etal-2025-pbi} jailbreaks vision-language models by combining image and text perturbations in a black-box setting, showing that cross-modal systems can be attacked through interactions between inputs rather than through any single prompt.

\paragraph{Practical Defenses and Evidence Checks} Several papers moved from finding failures to making outputs easier to check. Self-refine with formatting was proposed as a training-free jailbreak defense \citep{kim-cho-2025-break} that reduced attack success rates even on models without safety tuning, though the authors note it may need several rounds and does not work for every model. For claim verification, Minimal Evidence Groups identify the smallest non-redundant evidence set that fully supports a claim \citep{li-etal-2025-minimal} and outperform single-step retrieval. Together, these papers make the 2025 direction concrete: trust is not just about refusing bad requests or giving fluent answers; it is about having a checkable reason for the system's behavior.

\subsection{Phase 5: Mechanistic Trust \& Safety at Scale (2026)}

The sixth edition is the largest to date (41 papers) and marks two shifts: (i) explainability resurges via mechanistic interpretability methods that probe model internals rather than generating post-hoc explanations, and (ii) Machine Ethics \& Safety reaches its peak as the community confronts alignment fragility at scale. Three themes are emphasized below: mechanistic understanding of model behavior, safety alignment and its limits, and truthfulness in complex contexts.

\paragraph{Mechanistic Interpretability} The most striking development in 2026 is the return of explainability research (11 papers, up from 2 in 2025), driven by a fundamentally different methodology than the post-hoc attribution methods of 2021--2022. Linear probes reveal that high classification accuracy in distinguishing reasoning types reflects task format confounds rather than genuine computational differences in model hidden states~\citep{sahoo-etal-2026-linear}. Sparse autoencoders trained on multilingual data enable principled layer selection for activation steering across languages~\citep{al-ghussin-etal-2026-multilingual}. Single-layer activation edits can easily corrupt factual recall but rarely repair it, revealing a fundamental asymmetry in how knowledge is stored~\citep{bugaud-2026-single}. Concept-tracking probes allow monitoring what a model is ``thinking'' during normal operation~\citep{abdelwahab-etal-2026-thinking}. Rather than asking \emph{what} a model attends to, these works ask \emph{what happens when we intervene on specific internal components}.

\paragraph{Safety Alignment and Its Limits} Machine Ethics \& Safety reaches its highest count (8 papers), with a focus on the fragility of current alignment. \citet{bakman-etal-2026-hair} prove that static black-box evaluation cannot guarantee post-update alignment: models can pass all safety tests yet become severely misaligned after a single benign gradient update, with the capacity for latent adversarial behavior growing with model scale. The geometry of refusal is shown to be linearly manipulable---a steerable ``safety axis'' that serves as both vulnerability and defense primitive~\citep{ratnakar-vats-2026-geometry}. Overrefusal is traced to non-harmful linguistic cues in training data that models learn to associate with refusal~\citep{xue-etal-2026-deactivating}. Safety behavior proves domain-dependent, with compliance rates varying from 15\% (human trafficking) to 86\% (surveillance design) across ethical domains~\citep{bugaud-2026-domain}. The emerging lesson is that safety is not a scalar property but a domain-conditional, and geometrically localizable feature of model representations.

\paragraph{Truthfulness in Complex Contexts} Truthfulness remains the most represented dimension (13 papers), but the problems grow more subtle. Ghost Context~\citep{namboothiri-2026-ghost} formalizes a new failure mode---misattributed grounding---where models use evidence from the wrong part of a long context, producing errors invisible to standard faithfulness metrics. Quantization's impact on factual knowledge recall is characterized across methods and model scales~\citep{wang-etal-2026-compressed}. Prospective memory failures show that models drop formatting constraints under cognitive load~\citep{mittal-2026-forget}. In this edition we saw focus where truthfulness research moves from detecting hallucinations to understanding \emph{why} and \emph{where} factual knowledge degrades within model architectures.

\section{Structural Insights and Unexplored Directions}

Our longitudinal synthesis reveals both recurring structural patterns and persistent blind spots in the TrustNLP proceedings. In this section, we first distill four structural insights that emerge from the six-year arc of the workshop.
%—concerning the reactive nature of trust research, its dependence on interaction paradigms, the gap between output-level and internal-level evaluation, and the absence of a unifying theoretical framework. 
We then identify underexplored directions that, taken together with these insights, delineate the most pressing opportunities for the research community. The insights and gaps are deeply interrelated. The structural patterns explain why certain directions remain underexplored, while the gaps themselves constrain the field's ability to move from documenting trust failures to engineering trust guarantees.

\subsection{Structural Insights}

\paragraph{Insight 1: Trust Topics trail Capability Events.}
We observe that shifts in TrustNLP's topical focus follow notable events in the broader AI field. The release of the first high-impact chat models (late 2022) preceded the simultaneous activation of Robustness, Truthfulness, and Privacy in the 2023 edition---dimensions that had been absent or minimal in prior years. The availability of frontier models and open-weight alternatives (2023--2024) preceded the emergence of multi-dimensional trade-off research in 2024, where papers began studying conflicts between fairness, explainability, and performance. The rise of agentic and multimodal systems (2024--2025) preceded the 2025--2026 focus on safety alignment and mechanistic interpretability. In each case, the workshop edition following a capability event showed heightened attention to trust challenges that the new capability made salient. It is natural that the community's research agenda would respond to newly observable failure modes.

\paragraph{Insight 2: Trust Evidence Depends on Interaction Mode.}
What counts as evidence of trustworthiness has varied with how humans interact with models. Under direct model access (2021--2022), trust evidence meant feature attributions and rationale extraction~\cite{tang-surdeanu-2021-interpretability}. Under API-mediated access (2023--2024), it shifted to factuality scores and calibration~\cite{chern-etal-2023-improving}. Under agentic and multi-step use (2025), it became trajectory-level safety guarantees~\cite{rabinovich-anaby-tavor-2025-robustness}. However, this progression is not purely linear. The 2026 resurgence of explainability (2 papers in 2025 to 13 in 2026) shows that the field can revisit earlier trust questions with new methods---mechanistic interpretability (sparse autoencoders, activation steering, linear probes) addresses the same ``why did the model do that?'' question as 2021-era attribution methods, but through causal intervention rather than post-hoc correlation. The interaction paradigm shapes \emph{which} trust questions are most urgent, but earlier questions do not disappear. They return when new methodology makes them tractable again.

\paragraph{Insight 3: Outputs $\neq$ Internals.}
Output-level audits, the dominant evaluation paradigm across all six editions, systematically underestimate latent model behaviors. This pattern is visible directly within our corpus: fairness audits based on output distributions miss internal representational biases detectable only through probing \cite{balkir-etal-2022-challenges}; reliability benchmarks (\S4.2) missed contamination-driven inflation of benchmark scores \cite{aiyappa-etal-2023-trust}; adversarial evaluations (\S4.3) missed latent attack surfaces discoverable only through automated red-teaming \cite{lal-etal-2024-automated}; and function-calling evaluations (\S4.4) missed brittleness to toolkit expansion \cite{rabinovich-anaby-tavor-2025-robustness}. The broader literature outside TrustNLP corroborates this gap at a mechanistic level: \citet{cao-etal-2022-intrinsic} show a lack of alignment between fairness metrics computed on internal representations and on model outputs; the KEEN probe predicts model factuality directly from internal representations without generating any text \cite{gottesman-geva-2024-estimating}; and parametric-knowledge-trace probing reveals that unlearning methods minimally alter concept vectors and primarily suppress them at inference time, leaving the underlying knowledge intact and recoverable \cite{hong-etal-2025-intrinsic}. The implication is architectural: trustworthy evaluation must incorporate internal probes, not just behavioral tests, as a first-class component, particularly as agentic systems make autonomous decisions based on latent representations that never surface in user-visible outputs.

\begin{comment}
\paragraph{Insight 4: The Need for a Unified Framework.}
Despite 144 papers across six editions, TrustNLP lacks a unifying theoretical framework that connects its constituent concerns, such as fairness, robustness, factuality, privacy, and calibration, into a coherent whole. Each phase introduced new trust dimensions without formally relating them to prior ones: interpretability (\S4.1) was not reconciled with reliability (\S4.2), which was not reconciled with controllability (\S4.3--4.4). Existing external frameworks such as DecodingTrust \cite{wang-etal-2024-decodingtrust} and TrustLLM \cite{huang2024trustllm} enumerate dimensions but do not formalize their interactions or trade-offs. Yet the proceedings themselves reveal deep interdependencies: fairness conflicts with efficiency \cite{bui-von-der-wense-2024-trade} and explainability~\citep{brandl-etal-2024-interplay}; robustness conflicts with calibration; and privacy conflicts with utility. A unified framework would need to (i) define trust as a multi-objective optimization problem with explicit trade-off surfaces, (ii) specify how trust requirements change as a function of the interaction paradigm (Insight~2), and (iii) incorporate both output-level and internal-level evidence (Insight~3). Without such a framework, the field risks fragmentation: each new capability emergence produces yet another disconnected line of trust research rather than cumulative theoretical progress.
\end{comment}

\paragraph{Insight 4: The Need for a Unified Framework.}
Despite 144 papers across six editions, TrustNLP lacks a unifying theoretical framework that connects its constituent concerns, such as fairness, robustness, factuality, privacy, and calibration. Existing external frameworks such as DecodingTrust \cite{wang-etal-2024-decodingtrust} and TrustLLM \cite{huang2024trustllm} enumerate dimensions but do not formalize their interactions or trade-offs. However, the proceedings themselves reveal deep interdependencies among these dimensions: adapter modules that improve performance and efficiency can come at the cost of fairness \cite{bui-von-der-wense-2024-trade}, and improving fairness does not reliably improve explainability, or vice versa \cite{brandl-etal-2024-interplay}. A unified framework would need to explicitly account for these trade-offs rather than treating each dimension in isolation, specify how trust requirements change with the interaction paradigm, and incorporate both output-level and internal-level evidence. Without such a framework, the field risks fragmentation: each new capability emergence produces yet another disconnected line of trust research.

%\subsection{Gaps and Underexplored Directions}

%Two persistent gaps stand out from our analysis. First, despite trust being fundamentally a human judgment, only 8 of 144 papers involve user studies or measure perceived trust. The proceedings overwhelmingly evaluate trust through automated metrics rather than human experience. As models move toward agentic, long-term interactions (\S4.4), understanding how users calibrate trust over time---and how that calibration breaks down---becomes essential. Future work should integrate insights from HCI and cognitive science on trust formation, repair, and erosion \cite{ehsan-etal-2020-hcxai}. Second, 

In addition to above insights, we observe that the rapid growth in evaluation papers has produced benchmarks that saturate quickly, with several works already questioning existing metrics \cite{hosseini-etal-2023-empirical}. The field needs dynamic, adversarially-maintained benchmarks that evolve with model capabilities, rather than static test sets that become training data for the next generation. Benchmark design should also account for the interaction paradigm shifts identified in Insight~2. For example, a benchmark designed for single-turn generation is inadequate for evaluating multi-turn agentic behavior.

\section{Conclusion}

\begin{comment}
Six years of TrustNLP proceedings reveal a field shaped more by external capability emergence than by internal theoretical development. Each phase, such as interpretability, reliability, controllability, and now agentic safety, was catalyzed by deployment events rather than anticipated by the research community. The 6--9 month reactive lag (Insight~1), the paradigm-dependent nature of trust definitions (Insight~2), the persistent gap between output-level evaluation and internal model behavior (Insight~3), and the absence of a unifying framework (Insight~4) are not independent observations; they are symptoms of a field that has grown rapidly without consolidating its foundations.

Three priorities emerge for future editions. First, \emph{proactive engagement}: trust research should be embedded within capability development pipelines, not conducted post-deployment. Second, \emph{internal auditing}: evaluation must move beyond behavioral testing to mechanistic verification of what models know, retain, and can be adversarially induced to express. Third, \emph{theoretical unification}: the community needs formal frameworks that relate fairness, robustness, calibration, and privacy as interacting constraints rather than independent checklists.

The growth from 9 to 39 papers per edition demonstrates sustained community investment. Whether that investment yields cumulative progress or fragmented reactions to the next capability emergence depends on whether the field can transition from documenting trust failures to engineering trust guarantees.
\end{comment}

Across six editions and 144 papers, TrustNLP shows a field shaped more by external capability emergence than by internal theoretical development: trust paradigms have shifted from interpretability to reliability to controllability in lockstep with the dominant human--AI interaction mode, with a stable multi-month research lag behind each shock. Several findings go deeper. E.g., output-level audits systematically miss latent model behaviors, so trustworthy evaluation must incorporate internal probes as a first-class component. Second, the field's trust dimensions, such as fairness, robustness, factuality, privacy, and calibration, interact in ways that no single benchmark currently captures. Third, the persistent absence of a unifying framework means that each new capability emergence produces another disconnected line of trust research rather than cumulative theoretical progress.

The growth from 8 to 41 archival papers per edition demonstrates sustained community investment. But documenting failures is not the same as preventing them. The field must now move beyond observation. Research should proactively identify modes when models must not fail, then verify mechanistically and not just behaviorally.

\section{Limitations}

Our analysis is bounded by several scope factors that readers should consider when generalizing the findings. First, our analysis covers a single workshop venue (TrustNLP) over six editions; while we argue this venue is representative of the ACL community's trust agenda, our findings may not generalize to trustworthy-AI work published at FAccT, AIES, ICLR, NeurIPS, or non-archival venues. Second, our synthesis is restricted to archival proceedings papers; non-archival presentations, posters, and tutorial content are excluded. Third, the paper is anglocentric, the cited literature, the lexical analysis, and the example paradigms are drawn primarily from English-language NLP. The fact that multilingual safety remains underexplored is itself a symptom of this bias in the source corpus. Fourth, we identify structural patterns retrospectively. The reactive delay (Insight~1) is an observation across four capability emergence; whether this regularity holds for future shocks is an empirical question we cannot answer. Finally, this paper does not make an experimental contribution. Our claims are descriptive and analytic, drawing on the proceedings record rather than new measurements. Readers seeking benchmark numbers, model evaluations, or method comparisons should consult the individual papers cited.

\bibliography{custom}

\begin{thebibliography}{67}
\providecommand{\natexlab}[1]{#1}

\bibitem[{Abdelwahab et~al.(2026)Abdelwahab, Collins, Chen, Zhao, Mahmood, Zhu, Ali, and Rose}]{abdelwahab-etal-2026-thinking}
Mohamed Abdelwahab, Michelle~Yu Collins, Sihan Chen, Yi~Cheng Zhao, Zafarullah Mahmood, Jiading Zhu, Soliman Ali, and Jonathan Rose. 2026.
\newblock \href {https://doi.org/10.18653/v1/2026.trustnlp-main.9} {What are they thinking? delineation, probing, and tracking of concepts in {LLM}s}.
\newblock In \emph{Proceedings of the 6th Workshop on Trustworthy {NLP} ({T}rust{NLP} 2026)}, pages 121--179, San Diego, California. Association for Computational Linguistics.

\bibitem[{Adilazuarda(2024)}]{adilazuarda-2024-beyond}
Muhammad Adilazuarda. 2024.
\newblock \href {https://aclanthology.org/2024.trustnlp-1.1/} {Beyond {T}uring: A comparative analysis of approaches for detecting machine-generated text}.
\newblock In \emph{Proceedings of the 4th Workshop on Trustworthy Natural Language Processing (TrustNLP 2024)}, Mexico City, Mexico. Association for Computational Linguistics.

\bibitem[{Ahi and Valizadeh(2025)}]{ahi2025large}
Kiarash Ahi and Saeed Valizadeh. 2025.
\newblock \href {https://doi.org/10.1109/svcc65277.2025.11133642} {Large language models (llms) and generative ai in cybersecurity and privacy: A survey of dual-use risks, ai-generated malware, explainability, and defensive strategies}.
\newblock In \emph{2025 Silicon Valley Cybersecurity Conference (SVCC)}, pages 1--8. IEEE.

\bibitem[{Aiyappa et~al.(2023)Aiyappa, An, Kwak, and Ahn}]{aiyappa-etal-2023-trust}
Rachith Aiyappa, Jisun An, Haewoon Kwak, and Yong-Yeol Ahn. 2023.
\newblock \href {https://doi.org/10.18653/v1/2023.trustnlp-1.5} {Can we trust the evaluation on {C}hat{GPT}?}
\newblock In \emph{Proceedings of the 3rd Workshop on Trustworthy Natural Language Processing (TrustNLP 2023)}, pages 47--54, Toronto, Canada. Association for Computational Linguistics.

\bibitem[{Al~Ghussin et~al.(2026)Al~Ghussin, Gurgurov, Baeumel, van Genabith, Schramowski, and Ostermann}]{al-ghussin-etal-2026-multilingual}
Yusser Al~Ghussin, Daniil Gurgurov, Tanja Baeumel, Josef van Genabith, Patrick Schramowski, and Simon Ostermann. 2026.
\newblock \href {https://doi.org/10.18653/v1/2026.trustnlp-main.24} {Multilingual steering by design: Multilingual sparse autoencoders and principled layer selection}.
\newblock In \emph{Proceedings of the 6th Workshop on Trustworthy {NLP} ({T}rust{NLP} 2026)}, pages 364--401, San Diego, California. Association for Computational Linguistics.

\bibitem[{Anthropic(2023)}]{anthropic2023claude2}
Anthropic. 2023.
\newblock \href {https://www-cdn.anthropic.com/bd2a28d2535bfb0494cc8e2a3bf135d2e7523226/Model-Card-Claude-2.pdf} {Model card and evaluations for claude models}.
\newblock Anthropic Model Card.

\bibitem[{Arnold et~al.(2023)Arnold, Yesilbas, and Weinzierl}]{arnold-etal-2023-driving}
Stefan Arnold, Dilara Yesilbas, and Sven Weinzierl. 2023.
\newblock \href {https://doi.org/10.18653/v1/2023.trustnlp-1.2} {Driving context into text-to-text privatization}.
\newblock In \emph{Proceedings of the 3rd Workshop on Trustworthy Natural Language Processing (TrustNLP 2023)}, pages 15--25, Toronto, Canada. Association for Computational Linguistics.

\bibitem[{Azarpanah and Farhadloo(2021)}]{azarpanah-farhadloo-2021-measuring}
Hossein Azarpanah and Mohsen Farhadloo. 2021.
\newblock \href {https://doi.org/10.18653/v1/2021.trustnlp-1.2} {Measuring biases of word embeddings: What similarity measures and descriptive statistics to use?}
\newblock In \emph{Proceedings of the First Workshop on Trustworthy Natural Language Processing}, pages 8--14, Online. Association for Computational Linguistics.

\bibitem[{Bakman et~al.(2026)Bakman, Yaldiz, Avestimehr, and Karimireddy}]{bakman-etal-2026-hair}
Yavuz~Faruk Bakman, Duygu~Nur Yaldiz, Salman Avestimehr, and Sai~Praneeth Karimireddy. 2026.
\newblock \href {https://doi.org/10.18653/v1/2026.trustnlp-main.10} {Hair-trigger alignment: Black-box evaluation cannot guarantee post-update alignment}.
\newblock In \emph{Proceedings of the 6th Workshop on Trustworthy {NLP} ({T}rust{NLP} 2026)}, pages 180--203, San Diego, California. Association for Computational Linguistics.

\bibitem[{Balkir et~al.(2022)Balkir, Kiritchenko, Nejadgholi, and Fraser}]{balkir-etal-2022-challenges}
Esma Balkir, Svetlana Kiritchenko, Isar Nejadgholi, and Kathleen Fraser. 2022.
\newblock \href {https://aclanthology.org/2022.trustnlp-1.8/} {Challenges in applying explainability methods to improve the fairness of {NLP} models}.
\newblock In \emph{Proceedings of the 2nd Workshop on Trustworthy Natural Language Processing (TrustNLP 2022)}, pages 80--92, Seattle, U.S.A. Association for Computational Linguistics.

\bibitem[{Brandl et~al.(2024)Brandl, Bugliarello, and Chalkidis}]{brandl-etal-2024-interplay}
Stephanie Brandl, Emanuele Bugliarello, and Ilias Chalkidis. 2024.
\newblock \href {https://aclanthology.org/2024.trustnlp-1.10/} {On the interplay between fairness and explainability}.
\newblock In \emph{Proceedings of the 4th Workshop on Trustworthy Natural Language Processing (TrustNLP 2024)}, pages 94--108, Mexico City, Mexico. Association for Computational Linguistics.

\bibitem[{Bugaud(2026{\natexlab{a}})}]{bugaud-2026-domain}
Zacharie Bugaud. 2026{\natexlab{a}}.
\newblock \href {https://doi.org/10.18653/v1/2026.trustnlp-main.42} {Domain-dependent safety behavior in open-weight {LLM}s: An empirical study across seven ethical domains}.
\newblock In \emph{Proceedings of the 6th Workshop on Trustworthy {NLP} ({T}rust{NLP} 2026)}, pages 557--562, San Diego, California. Association for Computational Linguistics.

\bibitem[{Bugaud(2026{\natexlab{b}})}]{bugaud-2026-single}
Zacharie Bugaud. 2026{\natexlab{b}}.
\newblock \href {https://doi.org/10.18653/v1/2026.trustnlp-main.38} {Single-layer activation edits easily corrupt factual recall but rarely repair it}.
\newblock In \emph{Proceedings of the 6th Workshop on Trustworthy {NLP} ({T}rust{NLP} 2026)}, pages 515--527, San Diego, California. Association for Computational Linguistics.

\bibitem[{Bui and Von Der~Wense(2024)}]{bui-von-der-wense-2024-trade}
Minh~Duc Bui and Katharina Von Der~Wense. 2024.
\newblock \href {https://aclanthology.org/2024.trustnlp-1.4/} {The trade-off between performance, efficiency, and fairness in adapter modules for text classification}.
\newblock In \emph{Proceedings of the 4th Workshop on Trustworthy Natural Language Processing (TrustNLP 2024)}, Mexico City, Mexico. Association for Computational Linguistics.

\bibitem[{Cao et~al.(2025)Cao, Das, Kumarage, Wan, Krishna, Mehrabi, Dhamala, Ramakrishna, Galystan, Kumar, Gupta, and Chang}]{trustnlp-2025-trustworthy}
Trista Cao, Anubrata Das, Tharindu Kumarage, Yixin Wan, Satyapriya Krishna, Ninareh Mehrabi, Jwala Dhamala, Anil Ramakrishna, Aram Galystan, Anoop Kumar, Rahul Gupta, and Kai-Wei Chang, editors. 2025.
\newblock \href {https://aclanthology.org/2025.trustnlp-main.0/} {\emph{Proceedings of the 5th Workshop on Trustworthy NLP ({TrustNLP} 2025)}}. Association for Computational Linguistics, Albuquerque, New Mexico.

\bibitem[{Cao et~al.(2023)Cao, Dawa, Qun, and Nyima}]{cao-etal-2023-pay-attention}
Xi~Cao, Dolma Dawa, Nuo Qun, and Trashi Nyima. 2023.
\newblock \href {https://doi.org/10.18653/v1/2023.trustnlp-1.4} {Pay attention to the robustness of {C}hinese minority language models! syllable-level textual adversarial attack on {T}ibetan script}.
\newblock In \emph{Proceedings of the 3rd Workshop on Trustworthy Natural Language Processing (TrustNLP 2023)}, pages 35--46, Toronto, Canada. Association for Computational Linguistics.

\bibitem[{Cao et~al.(2022)Cao, Pruksachatkun, Chang, Gupta, Kumar, Dhamala, and Galstyan}]{cao-etal-2022-intrinsic}
Yang~Trista Cao, Yada Pruksachatkun, Kai-Wei Chang, Rahul Gupta, Varun Kumar, Jwala Dhamala, and Aram Galstyan. 2022.
\newblock \href {https://doi.org/10.18653/v1/2022.acl-short.62} {On the intrinsic and extrinsic fairness evaluation metrics for contextualized language representations}.
\newblock In \emph{Proceedings of the 60th Annual Meeting of the Association for Computational Linguistics (Volume 2: Short Papers)}, pages 561--570, Dublin, Ireland. Association for Computational Linguistics.

\bibitem[{Chen et~al.(2025)Chen, Chen, Chen, and Sra}]{chen2025standard}
Zichen Chen, Jiaao Chen, Jianda Chen, and Misha Sra. 2025.
\newblock \href {https://arxiv.org/abs/2502.15865} {Standard benchmarks fail--auditing llm agents in finance must prioritize risk}.
\newblock \emph{arXiv preprint arXiv:2502.15865}.

\bibitem[{Cheng et~al.(2025{\natexlab{a}})Cheng, Ding, Cao, Duan, Jia, Yuan, Wang, and Jia}]{cheng-etal-2025-pbi}
Ruoxi Cheng, Yizhong Ding, Shuirong Cao, Ranjie Duan, Xiaoshuang Jia, Shaowei Yuan, Zhiqiang Wang, and Xiaojun Jia. 2025{\natexlab{a}}.
\newblock \href {https://doi.org/10.18653/v1/2025.trustnlp-main.3} {{PBI}-attack: Prior-guided bimodal interactive black-box jailbreak attack for toxicity maximization}.
\newblock In \emph{Proceedings of the 5th Workshop on Trustworthy NLP (TrustNLP 2025)}, pages 23--40, Albuquerque, New Mexico. Association for Computational Linguistics.

\bibitem[{Cheng et~al.(2025{\natexlab{b}})Cheng, Ding, Cao, Wang, and Shao}]{cheng-etal-2025-gibberish}
Ruoxi Cheng, Yizhong Ding, Shuirong Cao, Zhiqiang Wang, and Shitong Shao. 2025{\natexlab{b}}.
\newblock \href {https://doi.org/10.18653/v1/2025.trustnlp-main.2} {Gibberish is all you need for membership inference detection in contrastive language-audio pretraining}.
\newblock In \emph{Proceedings of the 5th Workshop on Trustworthy NLP (TrustNLP 2025)}, pages 13--22, Albuquerque, New Mexico. Association for Computational Linguistics.

\bibitem[{Chern et~al.(2023)Chern, Wang, Das, Sharma, Liu, and Neubig}]{chern-etal-2023-improving}
I-chun Chern, Zhiruo Wang, Sanjan Das, Bhavuk Sharma, Pengfei Liu, and Graham Neubig. 2023.
\newblock \href {https://doi.org/10.18653/v1/2023.trustnlp-1.6} {Improving factuality of abstractive summarization via contrastive reward learning}.
\newblock In \emph{Proceedings of the 3rd Workshop on Trustworthy Natural Language Processing (TrustNLP 2023)}, pages 55--60, Toronto, Canada. Association for Computational Linguistics.

\bibitem[{Devlin et~al.(2019)Devlin, Chang, Lee, and Toutanova}]{devlin-etal-2019-bert}
Jacob Devlin, Ming-Wei Chang, Kenton Lee, and Kristina Toutanova. 2019.
\newblock \href {https://doi.org/10.18653/v1/N19-1423} {{BERT}: Pre-training of deep bidirectional transformers for language understanding}.
\newblock In \emph{Proceedings of the 2019 Conference of the North {A}merican Chapter of the Association for Computational Linguistics: Human Language Technologies, Volume 1 (Long and Short Papers)}, pages 4171--4186, Minneapolis, Minnesota. Association for Computational Linguistics.

\bibitem[{Fraser et~al.(2022)Fraser, Kiritchenko, and Balkir}]{fraser-etal-2022-moral}
Kathleen~C. Fraser, Svetlana Kiritchenko, and Esma Balkir. 2022.
\newblock \href {https://doi.org/10.18653/v1/2022.trustnlp-1.3} {Does moral code have a moral code? probing {D}elphi's moral philosophy}.
\newblock In \emph{Proceedings of the 2nd Workshop on Trustworthy Natural Language Processing (TrustNLP 2022)}, pages 26--42, Seattle, U.S.A. Association for Computational Linguistics.

\bibitem[{Fu et~al.(2024)Fu, Xiao, Chen, Li, Papalexakis, Chien, and Dong}]{fu-etal-2024-cross}
Yu~Fu, Wen Xiao, Jia Chen, Jiachen Li, Evangelos Papalexakis, Aichi Chien, and Yue Dong. 2024.
\newblock \href {https://doi.org/10.18653/v1/2024.trustnlp-1.9} {Cross-task defense: Instruction-tuning {LLM}s for content safety}.
\newblock In \emph{Proceedings of the 4th Workshop on Trustworthy Natural Language Processing (TrustNLP 2024)}, pages 85--93, Mexico City, Mexico. Association for Computational Linguistics.

\bibitem[{Gottesman and Geva(2024)}]{gottesman-geva-2024-estimating}
Daniela Gottesman and Mor Geva. 2024.
\newblock \href {https://doi.org/10.18653/v1/2024.emnlp-main.232} {Estimating knowledge in large language models without generating a single token}.
\newblock In \emph{Proceedings of the 2024 Conference on Empirical Methods in Natural Language Processing}, pages 3994--4019, Miami, Florida, USA. Association for Computational Linguistics.

\bibitem[{Harvey et~al.(2025)Harvey, Koenecke, and Kizilcec}]{harvey2025don}
Emma Harvey, Allison Koenecke, and Rene~F Kizilcec. 2025.
\newblock \href {https://doi.org/10.1145/3706598.3713210} {" don't forget the teachers": Towards an educator-centered understanding of harms from large language models in education}.
\newblock In \emph{Proceedings of the 2025 CHI Conference on Human Factors in Computing Systems}, pages 1--19.

\bibitem[{Hong et~al.(2025)Hong, Yu, Yang, Ravfogel, and Geva}]{hong-etal-2025-intrinsic}
Yihuai Hong, Lei Yu, Haiqin Yang, Shauli Ravfogel, and Mor Geva. 2025.
\newblock \href {https://doi.org/10.18653/v1/2025.emnlp-main.985} {Intrinsic test of unlearning using parametric knowledge traces}.
\newblock In \emph{Proceedings of the 2025 Conference on Empirical Methods in Natural Language Processing}, pages 19513--19535, Suzhou, China. Association for Computational Linguistics.

\bibitem[{Hosseini et~al.(2023)Hosseini, Palangi, and Awadallah}]{hosseini-etal-2023-empirical}
Saghar Hosseini, Hamid Palangi, and Ahmed~Hassan Awadallah. 2023.
\newblock \href {https://aclanthology.org/2023.trustnlp-1.11/} {An empirical study of metrics to measure representational harms in pre-trained language models}.
\newblock In \emph{Proceedings of the 3rd Workshop on Trustworthy Natural Language Processing (TrustNLP 2023)}, pages 121--134, Toronto, Canada. Association for Computational Linguistics.

\bibitem[{Hovy and Yang(2021)}]{hovy-yang-2021-importance}
Dirk Hovy and Diyi Yang. 2021.
\newblock \href {https://doi.org/10.18653/v1/2021.naacl-main.49} {The importance of modeling social factors of language: Theory and practice}.
\newblock In \emph{Proceedings of the 2021 Conference of the North American Chapter of the Association for Computational Linguistics: Human Language Technologies}, pages 588--602, Online. Association for Computational Linguistics.

\bibitem[{Huang et~al.(2024)Huang, Sun, Wang, Wu, Zhang, Li, Gao, Huang, Lyu, Zhang et~al.}]{huang2024trustllm}
Yue Huang, Lichao Sun, Haoran Wang, Siyuan Wu, Qihui Zhang, Yuan Li, Chujie Gao, Yixin Huang, Wenhan Lyu, Yixuan Zhang, and 1 others. 2024.
\newblock \href {https://arxiv.org/abs/2401.05561} {{T}rust{LLM}: Trustworthiness in large language models}.
\newblock \emph{arXiv preprint arXiv:2401.05561}.

\bibitem[{Jeoung et~al.(2023)Jeoung, Diesner, and Kilicoglu}]{jeoung-etal-2023-examining}
Sullam Jeoung, Jana Diesner, and Halil Kilicoglu. 2023.
\newblock \href {https://aclanthology.org/2023.trustnlp-1.7/} {Examining the causal impact of first names on language models: The case of social commonsense reasoning}.
\newblock In \emph{Proceedings of the 3rd Workshop on Trustworthy Natural Language Processing (TrustNLP 2023)}, pages 61--72, Toronto, Canada. Association for Computational Linguistics.

\bibitem[{Khatun and Brown(2023)}]{khatun-brown-2023-reliability}
Aisha Khatun and Daniel Brown. 2023.
\newblock \href {https://doi.org/10.18653/v1/2023.trustnlp-1.8} {Reliability check: An analysis of {GPT}-3's response to sensitive topics and prompt wording}.
\newblock In \emph{Proceedings of the 3rd Workshop on Trustworthy Natural Language Processing (TrustNLP 2023)}, pages 73--95, Toronto, Canada. Association for Computational Linguistics.

\bibitem[{Kim and Cho(2025)}]{kim-cho-2025-break}
Heegyu Kim and Hyunsouk Cho. 2025.
\newblock \href {https://aclanthology.org/2025.trustnlp-main.7/} {Break the breakout: Reinventing {LM} defense against jailbreak attacks with self-refine}.
\newblock In \emph{Proceedings of the 5th Workshop on Trustworthy NLP (TrustNLP 2025)}, Albuquerque, New Mexico. Association for Computational Linguistics.

\bibitem[{Lal et~al.(2024)Lal, Lahoti, Sinha, Qin, and Balashankar}]{lal-etal-2024-automated}
Yash~Kumar Lal, Preethi Lahoti, Aradhana Sinha, Yao Qin, and Ananth Balashankar. 2024.
\newblock \href {https://aclanthology.org/2024.trustnlp-1.2/} {Automated adversarial discovery for safety classifiers}.
\newblock In \emph{Proceedings of the 4th Workshop on Trustworthy Natural Language Processing (TrustNLP 2024)}, Mexico City, Mexico. Association for Computational Linguistics.

\bibitem[{Li et~al.(2022)Li, Ma, and Lin}]{li-etal-2022-encoder}
Minghan Li, Xueguang Ma, and Jimmy Lin. 2022.
\newblock \href {https://aclanthology.org/2022.trustnlp-1.1/} {An encoder attribution analysis for dense passage retriever in open-domain question answering}.
\newblock In \emph{Proceedings of the 2nd Workshop on Trustworthy Natural Language Processing (TrustNLP 2022)}, pages 1--11, Seattle, U.S.A. Association for Computational Linguistics.

\bibitem[{Li et~al.(2025)Li, Chen, Kapadia, Ouyang, and Zhang}]{li-etal-2025-minimal}
Xiangci Li, Sihao Chen, Rajvi Kapadia, Jessica Ouyang, and Fan Zhang. 2025.
\newblock \href {https://doi.org/10.18653/v1/2025.trustnlp-main.8} {Minimal evidence group identification for claim verification}.
\newblock In \emph{Proceedings of the 5th Workshop on Trustworthy NLP (TrustNLP 2025)}, pages 103--111, Albuquerque, New Mexico. Association for Computational Linguistics.

\bibitem[{Liu et~al.(2024)Liu, Yang, Lei, Shen, Wang, Wei, Chu, Qin, and Ren}]{liu2024survey}
Lei Liu, Xiaoyan Yang, Junchi Lei, Yue Shen, Jian Wang, Peng Wei, Zhixuan Chu, Zhan Qin, and Kui Ren. 2024.
\newblock \href {https://arxiv.org/abs/2406.03712} {A survey on medical large language models: Technology, application, trustworthiness, and future directions}.
\newblock \emph{arXiv preprint arXiv:2406.03712}.

\bibitem[{Liu et~al.(2023)Liu, Yao, Ton, Zhang, Guo, Cheng, Klochkov, Taufiq, and Li}]{liu-etal-2024-trustworthy-llms-survey}
Yang Liu, Yuanshun Yao, Jean-Francois Ton, Xiaoying Zhang, Ruocheng Guo, Hao Cheng, Yegor Klochkov, Muhammad~Faaiz Taufiq, and Hang Li. 2023.
\newblock \href {https://arxiv.org/abs/2308.05374} {Trustworthy {LLM}s: A survey and guideline for evaluating large language models' alignment}.
\newblock \emph{arXiv preprint arXiv:2308.05374}.

\bibitem[{Matthews et~al.(2021)Matthews, Grasso, Mahoney, Chen, Wali, Middleton, Njie, and Matthews}]{matthews-etal-2021-gender}
Abigail Matthews, Isabella Grasso, Christopher Mahoney, Yan Chen, Esma Wali, Thomas Middleton, Mariama Njie, and Jeanna Matthews. 2021.
\newblock \href {https://doi.org/10.18653/v1/2021.trustnlp-1.6} {Gender bias in natural language processing across human languages}.
\newblock In \emph{Proceedings of the First Workshop on Trustworthy Natural Language Processing}, pages 45--54, Online. Association for Computational Linguistics.

\bibitem[{Mittal(2026)}]{mittal-2026-forget}
Avni Mittal. 2026.
\newblock \href {https://doi.org/10.18653/v1/2026.trustnlp-main.33} {Did you forget what {I} asked? prospective memory failures in large language models}.
\newblock In \emph{Proceedings of the 6th Workshop on Trustworthy {NLP} ({T}rust{NLP} 2026)}, pages 471--488, San Diego, California. Association for Computational Linguistics.

\bibitem[{Namboothiri(2026)}]{namboothiri-2026-ghost}
Rohith Namboothiri. 2026.
\newblock \href {https://doi.org/10.18653/v1/2026.trustnlp-main.19} {Ghost context: Measuring cross-context interference in long-context language models}.
\newblock In \emph{Proceedings of the 6th Workshop on Trustworthy {NLP} ({T}rust{NLP} 2026)}, pages 316--329, San Diego, California. Association for Computational Linguistics.

\bibitem[{OpenAI(2023)}]{openai2023gpt4}
OpenAI. 2023.
\newblock \href {https://arxiv.org/abs/2303.08774} {Gpt-4 technical report}.
\newblock \emph{arXiv preprint arXiv:2303.08774}.

\bibitem[{Ovalle et~al.(2024)Ovalle, Chang, Cao, Mehrabi, Zhao, Galstyan, Dhamala, Kumar, and Gupta}]{trustnlp-2024-trustworthy}
Anaelia Ovalle, Kai-Wei Chang, Yang~Trista Cao, Ninareh Mehrabi, Jieyu Zhao, Aram Galstyan, Jwala Dhamala, Anoop Kumar, and Rahul Gupta, editors. 2024.
\newblock \href {https://aclanthology.org/2024.trustnlp-1.0/} {\emph{Proceedings of the 4th Workshop on Trustworthy Natural Language Processing ({TrustNLP} 2024)}}. Association for Computational Linguistics, Mexico City, Mexico.

\bibitem[{Ovalle et~al.(2023)Ovalle, Chang, Mehrabi, Pruksachatkun, Galystan, Dhamala, Verma, Cao, Kumar, and Gupta}]{trustnlp-2023-trustworthy}
Anaelia Ovalle, Kai-Wei Chang, Ninareh Mehrabi, Yada Pruksachatkun, Aram Galystan, Jwala Dhamala, Apurv Verma, Trista Cao, Anoop Kumar, and Rahul Gupta, editors. 2023.
\newblock \href {https://aclanthology.org/2023.trustnlp-1.0/} {\emph{Proceedings of the 3rd Workshop on Trustworthy Natural Language Processing ({TrustNLP} 2023)}}. Association for Computational Linguistics, Toronto, Canada.

\bibitem[{Pavlich et~al.(2025)Pavlich, Ebadi, Tarbell, Linares, Tan, Humphreys, Das, Ghandiparsi, Haley, George, Slavin, Choo, Dietrich, and Rios}]{pavlich-etal-2025-beyond}
Ryan Pavlich, Nima Ebadi, Richard Tarbell, Billy Linares, Adrian Tan, Rachael Humphreys, Jayanta Das, Rambod Ghandiparsi, Hannah Haley, Jerris George, Rocky Slavin, Kim-Kwang~Raymond Choo, Glenn Dietrich, and Anthony Rios. 2025.
\newblock \href {https://doi.org/10.18653/v1/2025.trustnlp-main.1} {Beyond text-to-{SQL} for {IoT} defense: A comprehensive framework for querying and classifying {IoT} threats}.
\newblock In \emph{Proceedings of the 5th Workshop on Trustworthy NLP (TrustNLP 2025)}, pages 1--12, Albuquerque, New Mexico. Association for Computational Linguistics.

\bibitem[{Portillo~Wightman et~al.(2023)Portillo~Wightman, Delucia, and Dredze}]{portillo-wightman-etal-2023-strength}
Gwenyth Portillo~Wightman, Alexandra Delucia, and Mark Dredze. 2023.
\newblock \href {https://doi.org/10.18653/v1/2023.trustnlp-1.28} {Strength in numbers: Estimating confidence of large language models by prompt agreement}.
\newblock In \emph{Proceedings of the 3rd Workshop on Trustworthy Natural Language Processing (TrustNLP 2023)}, pages 326--362, Toronto, Canada. Association for Computational Linguistics.

\bibitem[{Pruksachatkun et~al.(2021)Pruksachatkun, Ramakrishna, Chang, Krishna, Dhamala, Guha, and Ren}]{trustnlp-2021-trustworthy}
Yada Pruksachatkun, Anil Ramakrishna, Kai-Wei Chang, Satyapriya Krishna, Jwala Dhamala, Tanaya Guha, and Xiang Ren, editors. 2021.
\newblock \href {https://aclanthology.org/2021.trustnlp-1.0/} {\emph{Proceedings of the First Workshop on Trustworthy Natural Language Processing}}. Association for Computational Linguistics, Online.

\bibitem[{Rabinovich and Anaby~Tavor(2025)}]{rabinovich-anaby-tavor-2025-robustness}
Ella Rabinovich and Ateret Anaby~Tavor. 2025.
\newblock \href {https://doi.org/10.18653/v1/2025.trustnlp-main.20} {On the robustness of agentic function calling}.
\newblock In \emph{Proceedings of the 5th Workshop on Trustworthy NLP (TrustNLP 2025)}, pages 298--304, Albuquerque, New Mexico. Association for Computational Linguistics.

\bibitem[{Raina and Gales(2023)}]{raina-gales-2023-sample}
Vyas Raina and Mark Gales. 2023.
\newblock \href {https://doi.org/10.18653/v1/2023.trustnlp-1.9} {Sample attackability in natural language adversarial attacks}.
\newblock In \emph{Proceedings of the 3rd Workshop on Trustworthy Natural Language Processing (TrustNLP 2023)}, pages 96--107, Toronto, Canada. Association for Computational Linguistics.

\bibitem[{Ratnakar and Vats(2026)}]{ratnakar-vats-2026-geometry}
Shivam Ratnakar and Kartikeya Vats. 2026.
\newblock \href {https://doi.org/10.18653/v1/2026.trustnlp-main.51} {The geometry of refusal: Linear instability in safety-aligned {LLM}s}.
\newblock In \emph{Proceedings of the 6th Workshop on Trustworthy {NLP} ({T}rust{NLP} 2026)}, pages 653--662, San Diego, California. Association for Computational Linguistics.

\bibitem[{Rawte et~al.(2025)Rawte, Tonmoy, Nag, Chadha, Sheth, and Das}]{rawte-etal-2025-factoid}
Vipula Rawte, S.m Towhidul~Islam Tonmoy, Shravani Nag, Aman Chadha, Amit Sheth, and Amitava Das. 2025.
\newblock \href {https://doi.org/10.18653/v1/2025.trustnlp-main.38} {{FACTOID}: {FAC}tual en{T}ailment f{O}r halluc{I}nation detection}.
\newblock In \emph{Proceedings of the 5th Workshop on Trustworthy NLP (TrustNLP 2025)}, pages 599--617, Albuquerque, New Mexico. Association for Computational Linguistics.

\bibitem[{Roberts et~al.(2021)Roberts, Driggs, Thorpe, Gilbey, Yeung, Ursprung, Aviles-Rivero, Etmann, McCague, Beer, Weir-McCall, Teng, Gkrania-Klotsas, AIX-COVNET, Rudd, Sala, and Sch{\"o}nlieb}]{roberts2021common}
Michael Roberts, Derek Driggs, Matthew Thorpe, Julian Gilbey, Michael Yeung, Stephan Ursprung, Angelica~I. Aviles-Rivero, Christian Etmann, Cathal McCague, Lucian Beer, Jonathan~R. Weir-McCall, Zhongzhao Teng, Effrossyni Gkrania-Klotsas, AIX-COVNET, James H.~F. Rudd, Evis Sala, and Carola-Bibiane Sch{\"o}nlieb. 2021.
\newblock \href {https://doi.org/10.1038/s42256-021-00307-0} {Common pitfalls and recommendations for using machine learning to detect and prognosticate for covid-19 using chest radiographs and ct scans}.
\newblock \emph{Nature Machine Intelligence}, 3(3):199--217.

\bibitem[{Sahoo et~al.(2026)Sahoo, Jain, Chadha, and Chaudhary}]{sahoo-etal-2026-linear}
Subramanyam Sahoo, Vinija Jain, Aman Chadha, and Divya Chaudhary. 2026.
\newblock \href {https://doi.org/10.18653/v1/2026.trustnlp-main.12} {Linear probes detect task format, not reasoning mode in language model hidden states}.
\newblock In \emph{Proceedings of the 6th Workshop on Trustworthy {NLP} ({T}rust{NLP} 2026)}, pages 227--239, San Diego, California. Association for Computational Linguistics.

\bibitem[{Shi et~al.(2025)Shi, Castellucci, Filice, Kuzi, Kravi, Agichtein, Rokhlenko, and Malmasi}]{shi-etal-2025-ambiguity}
Zhengyan Shi, Giuseppe Castellucci, Simone Filice, Saar Kuzi, Elad Kravi, Eugene Agichtein, Oleg Rokhlenko, and Shervin Malmasi. 2025.
\newblock \href {https://aclanthology.org/2025.trustnlp-main.4/} {Ambiguity detection and uncertainty calibration for question answering with large language models}.
\newblock In \emph{Proceedings of the 5th Workshop on Trustworthy NLP (TrustNLP 2025)}, Albuquerque, New Mexico. Association for Computational Linguistics.

\bibitem[{Subramani et~al.(2023)Subramani, Luccioni, Dodge, and Mitchell}]{subramani-etal-2023-detecting}
Nishant Subramani, Sasha Luccioni, Jesse Dodge, and Margaret Mitchell. 2023.
\newblock \href {https://doi.org/10.18653/v1/2023.trustnlp-1.18} {Detecting personal information in training corpora: an analysis}.
\newblock In \emph{Proceedings of the 3rd Workshop on Trustworthy Natural Language Processing (TrustNLP 2023)}, pages 208--220, Toronto, Canada. Association for Computational Linguistics.

\bibitem[{Tang and Surdeanu(2021)}]{tang-surdeanu-2021-interpretability}
Zheng Tang and Mihai Surdeanu. 2021.
\newblock \href {https://doi.org/10.18653/v1/2021.trustnlp-1.1} {Interpretability rules: Jointly bootstrapping a neural relation extractor with an explanation decoder}.
\newblock In \emph{Proceedings of the First Workshop on Trustworthy Natural Language Processing}, pages 1--7, Online. Association for Computational Linguistics.

\bibitem[{Touvron et~al.(2023)Touvron, Martin, Stone, Albert, Almahairi, Babaei, Bashlykov, Batra, Bhargava, Bhosale et~al.}]{touvron2023llama2}
Hugo Touvron, Louis Martin, Kevin Stone, Peter Albert, Amjad Almahairi, Yasmine Babaei, Nikolay Bashlykov, Soumya Batra, Prajjwal Bhargava, Shruti Bhosale, and 1 others. 2023.
\newblock \href {https://arxiv.org/abs/2307.09288} {Llama 2: Open foundation and fine-tuned chat models}.
\newblock \emph{arXiv preprint arXiv:2307.09288}.

\bibitem[{Venkit et~al.(2023)Venkit, Srinath, and Wilson}]{narayanan-venkit-etal-2023-automated}
Pranav~Narayanan Venkit, Mukund Srinath, and Shomir Wilson. 2023.
\newblock \href {https://doi.org/10.18653/v1/2023.trustnlp-1.3} {Automated ableism: An exploration of explicit disability biases in sentiment and toxicity analysis models}.
\newblock In \emph{Proceedings of the 3rd Workshop on Trustworthy Natural Language Processing (TrustNLP 2023)}, pages 26--34, Toronto, Canada. Association for Computational Linguistics.

\bibitem[{Verma et~al.(2022)Verma, Pruksachatkun, Chang, Galstyan, Dhamala, and Cao}]{trustnlp-2022-trustworthy}
Apurv Verma, Yada Pruksachatkun, Kai-Wei Chang, Aram Galstyan, Jwala Dhamala, and Yang~Trista Cao, editors. 2022.
\newblock \href {https://aclanthology.org/2022.trustnlp-1.0/} {\emph{Proceedings of the 2nd Workshop on Trustworthy Natural Language Processing ({TrustNLP} 2022)}}. Association for Computational Linguistics, Seattle, U.S.A.

\bibitem[{Wang et~al.(2018)Wang, Singh, Michael, Hill, Levy, and Bowman}]{wang-etal-2018-glue}
Alex Wang, Amanpreet Singh, Julian Michael, Felix Hill, Omer Levy, and Samuel~R. Bowman. 2018.
\newblock \href {https://doi.org/10.18653/v1/W18-5446} {{GLUE}: A multi-task benchmark and analysis platform for natural language understanding}.
\newblock In \emph{Proceedings of the 2018 {EMNLP} Workshop {B}lackbox{NLP}: Analyzing and Interpreting Neural Networks for {NLP}}, pages 353--355, Brussels, Belgium. Association for Computational Linguistics.

\bibitem[{Wang et~al.(2023)Wang, Chen, Pei, Xie, Kang, Zhang, Xu, Xiong, Dutta, Schaeffer et~al.}]{wang-etal-2024-decodingtrust}
Boxin Wang, Weixin Chen, Hengzhi Pei, Chulin Xie, Mintong Kang, Chenhui Zhang, Chejian Xu, Zidi Xiong, Ritik Dutta, Rylan Schaeffer, and 1 others. 2023.
\newblock \href {https://arxiv.org/abs/2306.11698} {{DecodingTrust}: A comprehensive assessment of trustworthiness in {GPT} models}.
\newblock In \emph{Advances in Neural Information Processing Systems 36: Annual Conference on Neural Information Processing Systems 2023, {NeurIPS} 2023 Datasets and Benchmarks Track}.

\bibitem[{Wang et~al.(2026)Wang, Wang, Feldhus, Ostermann, Cao, Schuetze, M{\"o}ller, and Schmitt}]{wang-etal-2026-compressed}
Qianli Wang, Mingyang Wang, Nils Feldhus, Simon Ostermann, Yuan Cao, Hinrich Schuetze, Sebastian M{\"o}ller, and Vera Schmitt. 2026.
\newblock \href {https://doi.org/10.18653/v1/2026.trustnlp-main.2} {Through a compressed lens: Investigating the impact of quantization on factual knowledge recall}.
\newblock In \emph{Proceedings of the 6th Workshop on Trustworthy {NLP} ({T}rust{NLP} 2026)}, pages 21--39, San Diego, California. Association for Computational Linguistics.

\bibitem[{Xue et~al.(2026)Xue, Qi, Liu, Chen, and Pedarsani}]{xue-etal-2026-deactivating}
Zhiyu Xue, Zimo Qi, Guangliang Liu, Bocheng Chen, and Ramtin Pedarsani. 2026.
\newblock \href {https://doi.org/10.18653/v1/2026.trustnlp-main.26} {Deactivating refusal triggers: Understanding and mitigating overrefusal in safety alignment}.
\newblock In \emph{Proceedings of the 6th Workshop on Trustworthy {NLP} ({T}rust{NLP} 2026)}, pages 402--412, San Diego, California. Association for Computational Linguistics.

\bibitem[{Yee et~al.(2023)Yee, Schoenauer~Sebag, Redfield, Eck, Sheng, and Belli}]{yee-etal-2023-keyword}
Kyra Yee, Alice Schoenauer~Sebag, Olivia Redfield, Matthias Eck, Emily Sheng, and Luca Belli. 2023.
\newblock \href {https://aclanthology.org/2023.trustnlp-1.10/} {A keyword based approach to understanding the overpenalization of marginalized groups by {E}nglish marginal abuse models on {T}witter}.
\newblock In \emph{Proceedings of the 3rd Workshop on Trustworthy Natural Language Processing (TrustNLP 2023)}, pages 108--120, Toronto, Canada. Association for Computational Linguistics.

\bibitem[{Yermilov et~al.(2023)Yermilov, Raheja, and Chernodub}]{yermilov-etal-2023-privacy}
Oleksandr Yermilov, Vipul Raheja, and Artem Chernodub. 2023.
\newblock \href {https://doi.org/10.18653/v1/2023.trustnlp-1.20} {Privacy- and utility-preserving {NLP} with anonymized data: A case study of pseudonymization}.
\newblock In \emph{Proceedings of the 3rd Workshop on Trustworthy Natural Language Processing (TrustNLP 2023)}, pages 232--241, Toronto, Canada. Association for Computational Linguistics.

\bibitem[{Yoo et~al.(2024)Yoo, Ahn, Song, and Kwak}]{yoo-etal-2024-exploring}
Kiyoon Yoo, Wonhyuk Ahn, Yeji Song, and Nojun Kwak. 2024.
\newblock \href {https://doi.org/10.18653/v1/2024.trustnlp-1.7} {Exploring causal mechanisms for machine text detection methods}.
\newblock In \emph{Proceedings of the 4th Workshop on Trustworthy Natural Language Processing (TrustNLP 2024)}, pages 71--78, Mexico City, Mexico. Association for Computational Linguistics.

\bibitem[{Ziems et~al.(2022)Ziems, Li, Zhang, and Yang}]{ziems-etal-2022-inducing}
Caleb Ziems, Minzhi Li, Anthony Zhang, and Diyi Yang. 2022.
\newblock \href {https://doi.org/10.18653/v1/2022.acl-long.257} {Inducing positive perspectives with text reframing}.
\newblock In \emph{Proceedings of the 60th Annual Meeting of the Association for Computational Linguistics (Volume 1: Long Papers)}, pages 3682--3700, Dublin, Ireland. Association for Computational Linguistics.

\end{thebibliography}

\appendix

\section{Annotation Instructions}
\label{sec:appendix-annotation-instructions}

The following instructions were provided to all annotators (human and LLM). Papers are classified based on their title and abstract only.

\paragraph{Trust Dimensions} (multi-label; mark each dimension that is a \emph{primary} focus of the paper):

\begin{itemize}
\item \textbf{Fairness \& Bias}: Papers whose primary focus is measuring, analyzing, or mitigating social bias (gender, race, disability, demographic, toxicity) in model outputs or representations.
\item \textbf{Robustness \& Adversarial}: Papers primarily proposing or evaluating adversarial attacks, defenses, backdoor/watermark detection, jailbreaks, red-teaming, or input perturbation robustness.
\item \textbf{Privacy}: Papers primarily addressing differential privacy, text anonymization/privatization, training data extraction, membership inference, PII protection, or machine unlearning.
\item \textbf{Machine Ethics \& Safety}: Papers primarily addressing alignment with human values, safety guardrails, harmful content generation, or moral reasoning in models.
\item \textbf{Truthfulness}: Papers primarily addressing hallucination, factuality, calibration, uncertainty estimation, or faithfulness of model outputs.
\item \textbf{Explainability}: Papers whose primary contribution is an interpretability method, explanation generation, feature attribution, rationale extraction, or probing of internal representations.
\end{itemize}

\section{LLM-Assisted Annotation Validation}
\label{sec:appendix-llm-validation}

To validate the manual annotations of 144 TrustNLP proceedings papers, we employed two LLMs from different model families as independent classifiers: Claude Sonnet~5 (Anthropic) and Nova Lite~2.0 (Amazon). Both models received identical prompts and were asked to classify each paper based solely on its \textbf{title and abstract} as published in the ACL Anthology. No full-text content was used.

\paragraph{Input Format.} Each paper was presented to the model as:

\begin{verbatim}
Title: {paper title}

Abstract: {paper abstract}

Classify this paper along both axes.
Respond with JSON only.
\end{verbatim}

\paragraph{System Prompt.} Both models received the following system prompt (reproduced verbatim):

\begin{quote}
\small
You are a strict research paper classifier. Given a paper's title and abstract, classify it along two axes.

\textbf{AXIS 1 --- Trust Dimensions} (multi-label, assign all that are a PRIMARY focus):

\begin{enumerate}
\item \textit{Fairness \& Bias}: Papers whose primary focus is measuring, analyzing, or mitigating social bias (gender, race, disability, demographic, toxicity) in model outputs or representations.
\item \textit{Robustness \& Adversarial}: Papers primarily proposing or evaluating adversarial attacks, defenses, backdoor/watermark detection, jailbreaks, red-teaming, or input perturbation robustness.
\item \textit{Privacy}: Papers primarily addressing differential privacy, text anonymization/privatization, training data extraction, membership inference, PII protection, or machine unlearning.
\item \textit{Machine Ethics \& Safety}: Papers primarily addressing alignment with human values, safety guardrails, harmful content generation, or moral reasoning in models.
\item \textit{Truthfulness}: Papers primarily addressing hallucination, factuality, calibration, uncertainty estimation, or faithfulness of model outputs.
\item \textit{Explainability}: Papers whose primary contribution is an interpretability method, explanation generation, feature attribution, rationale extraction, or probing of internal representations.
\end{enumerate}

IMPORTANT: Only assign a trust dimension if it is the PRIMARY focus. Most papers have 1--2 dimensions.
\end{quote}

\paragraph{Adjudication Process.} Papers where all three annotators (human + two LLMs) agreed required no further review. For disagreements, the human annotator reviewed the LLM suggestions and made a final decision, updating the label where the LLM rationale was judged more appropriate. This process yielded Cohen's $\kappa = 0.79$ (human vs.\ Sonnet~5) and $\kappa = 0.72$ (human vs.\ Nova Lite~2.0) on the trust dimension axis, indicating substantial agreement.

\paragraph{Models and Infrastructure.} Classification was performed via Amazon Bedrock with model identifiers \texttt{us.anthropic.claude-sonnet-5} and \texttt{us.amazon.nova-2-lite-v1:0}. Each model was constrained to 256 output tokens and asked to return structured JSON.

\section{Cross-Venue Comparison Methodology and Year-by-Year Analysis}
\label{sec:appendix-crossvenue}

To assess whether TrustNLP's topical trends generalize across the broader NLP community, we collected trust-related papers from four major *CL conferences: ACL, NAACL, EACL, and EMNLP (2021--2026). The classification pipeline consisted of two stages:

\paragraph{Stage 1: Keyword Filtering.} We scraped all main-track papers from ACL Anthology volumes for each venue and year. Papers were retained if their title contained at least one trust-related keyword from a curated list of 30 terms (including \textit{bias, fairness, robustness, adversarial, jailbreak, privacy, safety, alignment, hallucination, explainability, interpretability, toxicity, backdoor, unlearning}, among others). This produced a candidate set of 2,169 papers across all venues.

\paragraph{Stage 2: LLM Classification.} Each candidate paper's title and abstract were classified by Claude Sonnet~5 (via Amazon Bedrock) using the same trust dimension taxonomy and prompt structure as described in Appendix~\ref{sec:appendix-llm-validation}, yielding multi-label dimension assignments. Papers that failed classification (API errors) were excluded, resulting in 2,169 successfully classified papers across the four venues. To ensure methodological consistency, TrustNLP papers were also classified by Sonnet~5 using the same prompt, and the Sonnet labels (rather than human-adjudicated labels) are used for the cross-venue comparison reported below.

\begin{figure}[t]
\centering
\includegraphics[width=\columnwidth]{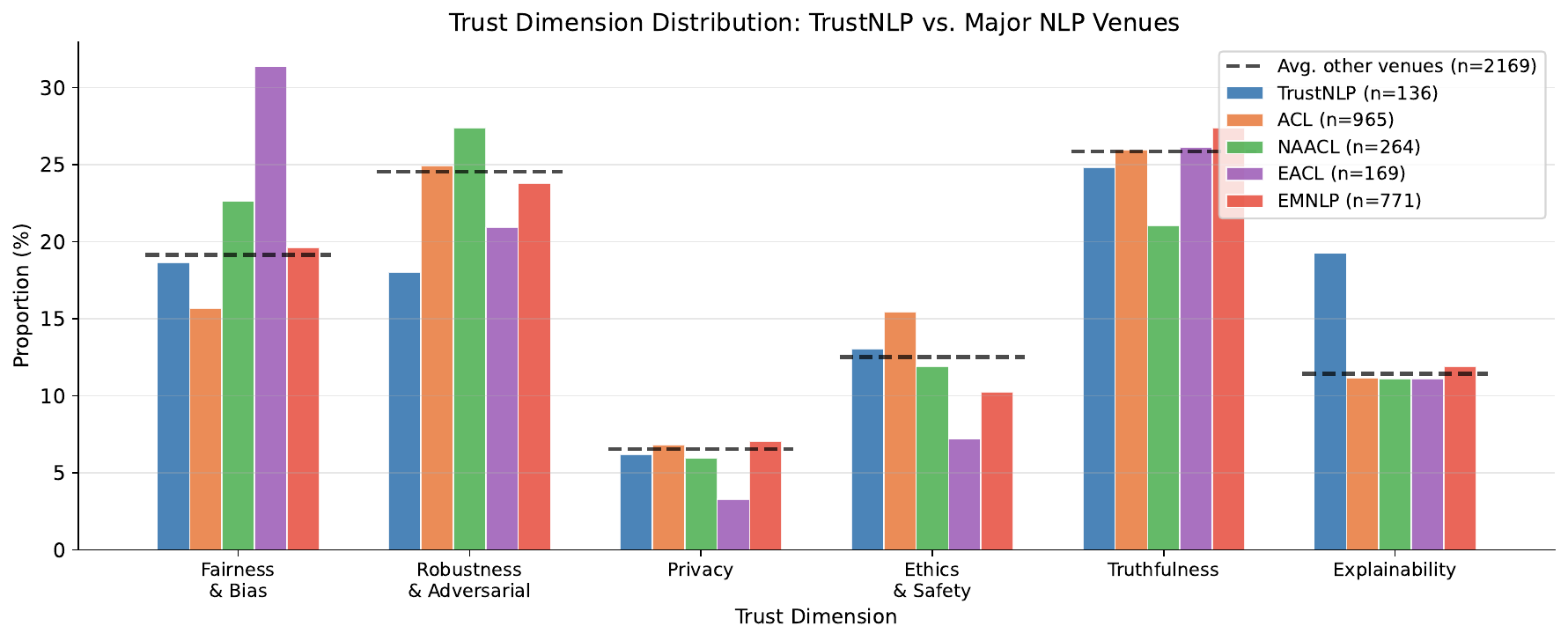}
\caption{Trust dimension distribution: TrustNLP vs.\ major NLP venues (aggregated across all available years). Dashed line shows the average across ACL, NAACL, EACL, and EMNLP.}
\label{fig:venue-comparison}
\end{figure}

\paragraph{Results.} Figure~\ref{fig:venue-comparison} shows the aggregate topical distribution across all years for each venue, with the dashed line indicating the average proportion across the four *CL conferences.
While we see some variation across conferences, TrustNLP closely traces the average representation of a topic each year across each broad category.

\onecolumn
\section{Complete Annotation Data}
\label{sec:appendix-annotation-review}

Table~\ref{tab:annotation-review} presents the complete annotation labels from the second review round for all 144 TrustNLP proceedings papers. Three independent annotators assigned trust dimension labels: one human expert (H), Claude Sonnet~5 (S), and Amazon Nova Lite~2.0 (N). Dimension abbreviations: F = Fairness \& Bias, R = Robustness \& Adversarial, P = Privacy, ES = Machine Ethics \& Safety, T = Truthfulness, E = Explainability.

{\tiny
\begin{longtable}{@{}cp{8.5cm}ccc@{}}
\caption{Complete annotation labels from the second review round. H = Human, S = Claude Sonnet~5, N = Nova Lite~2.0.}
\label{tab:annotation-review} \\
\toprule
\textbf{Year} & \textbf{Title} & \textbf{H} & \textbf{S} & \textbf{N} \\
\midrule
\endfirsthead
\toprule
\textbf{Year} & \textbf{Title} & \textbf{H} & \textbf{S} & \textbf{N} \\
\midrule
\endhead
\midrule
\multicolumn{5}{r}{\emph{Continued on next page}} \\
\endfoot
\bottomrule
\endlastfoot
2021 & Interpretability Rules: Jointly Bootstrapping a Neural Relation Extractorwith an Explanation Decoder & E & E & E \\
2021 & Measuring Biases of Word Embeddings: What Similarity Measures and Descriptive Statistics to Use? & F & F & F \\
2021 & Private Release of Text Embedding Vectors & P & P & P \\
2021 & Accountable Error Characterization & E & E & E \\
2021 & xER: An Explainable Model for Entity Resolution using an Efficient Solution for the Clique Partitioning Problem & E & E & E \\
2021 & Gender Bias in Natural Language Processing Across Human Languages & F & F & F \\
2021 & Interpreting Text Classifiers by Learning Context-sensitive Influence of Words & E & E & E \\
2021 & Towards Benchmarking the Utility of Explanations for Model Debugging & E & E & E \\
2022 & An Encoder Attribution Analysis for Dense Passage Retriever in Open-Domain Question Answering & --- & E & E \\
2022 & Attributing Fair Decisions with Attention Interventions & F, E & F, E & F, E \\
2022 & Does Moral Code have a Moral Code? Probing Delphi’s Moral Philosophy & ES & ES, E & F, ES, E \\
2022 & The Cycle of Trust and Responsibility in OutsourcedAI & ES & --- & ES \\
2022 & Explaining NeuralNLPModels for the Joint Analysis of Open-and-Closed-Ended Survey Answers & E & E & E \\
2022 & The Irrationality of Neural Rationale Models & E & E & E \\
2022 & An Empirical Study on Pseudo-log-likelihood Bias Measures for Masked Language Models Using Paraphrased Sentences & F & F & F \\
2022 & Challenges in Applying Explainability Methods to Improve the Fairness ofNLPModels & F, E & F, E & F, E \\
2023 & Towards Faithful Explanations for Text Classification with Robustness Improvement and Explanation Guided Training & E & R, E & R, T, E \\
2023 & A Keyword Based Approach to Understanding the Overpenalization of Marginalized Groups byEnglish Marginal Abuse Models onTwitter & F & F & F \\
2023 & An Empirical Study of Metrics to Measure Representational Harms in Pre-Trained Language Models & F & F & F \\
2023 & Linguistic Properties of Truthful Response & T & T & T \\
2023 & Debunking Biases in Attention & F & F & F \\
2023 & Guiding Text-to-Text Privatization by Syntax & P & P & P \\
2023 & Are fairness metric scores enough to assess discrimination biases in machine learning? & F & F & F \\
2023 & DEPTH+: An Enhanced Depth Metric forWikipedia Corpora Quality & --- & --- & --- \\
2023 & Distinguishing Fact from Fiction: A Benchmark Dataset for Identifying Machine-Generated Scientific Papers in theLLMEra. & T & T & T, E \\
2023 & Detecting Personal Information in Training Corpora: an Analysis & P & P & P \\
2023 & Enhancing textual counterfactual explanation intelligibility through Counterfactual Feature Importance & E & E & E \\
2023 & Driving Context into Text-to-Text Privatization & P & P & P \\
2023 & Privacy- and Utility-PreservingNLPwith Anonymized data: A case study of Pseudonymization & P & P & P \\
2023 & GPTs Don’t Keep Secrets: Searching for Backdoor Watermark Triggers in Autoregressive Language Models & R & R & R \\
2023 & Make Text Unlearnable: Exploiting Effective Patterns to Protect Personal Data & P & R, P & P \\
2023 & Training Data Extraction From Pre-trained Language Models: A Survey & R, P & P & P \\
2023 & Expanding Scope: AdaptingEnglish Adversarial Attacks toChinese & R & R & R \\
2023 & IMBERT: MakingBERTImmune to Insertion-based Backdoor Attacks & R & R & R \\
2023 & On The Real-world Performance of Machine Translation: Exploring Social Media Post-authors’ Perspectives & ES, T & --- & ES, T \\
2023 & Enabling Classifiers to Make Judgements Explicitly Aligned with Human Values & ES & --- & F, ES, E \\
2023 & Strength in Numbers: Estimating Confidence of Large Language Models by Prompt Agreement & T & T & T \\
2023 & Automated Ableism: An Exploration of Explicit Disability Biases in Sentiment and Toxicity Analysis Models & F & F & F \\
2023 & Pay Attention to the Robustness ofChinese Minority Language Models! Syllable-level Textual Adversarial Attack onTibetan Script & R & R & R \\
2023 & Can we trust the evaluation onChatGPT? & T & T & F \\
2023 & Improving Factuality of Abstractive Summarization via Contrastive Reward Learning & T & T & T \\
2023 & Examining the Causal Impact of First Names on Language Models: The Case of Social Commonsense Reasoning & F & F, E & F, E \\
2023 & Reliability Check: An Analysis ofGPT-3’s Response to Sensitive Topics and Prompt Wording & ES & T & R, ES, T \\
2023 & Sample Attackability in Natural Language Adversarial Attacks & R & R & R \\
2024 & BeyondTuring: A Comparative Analysis of Approaches for Detecting Machine-Generated Text & --- & --- & --- \\
2024 & On the Interplay between Fairness and Explainability & F, E & F, E & F, E \\
2024 & Holistic Evaluation of Large Language Models: Assessing Robustness, Accuracy, and Toxicity for Real-World Applications & F, R & F, R, T & F, R \\
2024 & HGOT: Hierarchical Graph of Thoughts for Retrieval-Augmented In-Context Learning in Factuality Evaluation & T & T & T \\
2024 & Overconfidence is Key: Verbalized Uncertainty Evaluation in Large Language and Vision-Language Models & T & T & T \\
2024 & Tweak to Trust: Assessing the Reliability of Summarization Metrics in Contact Centers via Perturbed Summaries & T & --- & R \\
2024 & Flatness-Aware Gradient Descent for Safe ConversationalAI & ES & ES & ES \\
2024 & IntroducingGenCeption for MultimodalLLMBenchmarking: You May Bypass Annotations & --- & T & --- \\
2024 & Semantic-Preserving Adversarial Example Attack againstBERT & R & R & R \\
2024 & Sandwich attack: Multi-language Mixture Adaptive Attack onLLMs & R & R, ES & R, ES \\
2024 & Masking Latent Gender Knowledge for Debiasing Image Captioning & F & F & F \\
2024 & Automated Adversarial Discovery for Safety Classifiers & R & R & F, R, ES \\
2024 & BELIEVE: Belief-Enhanced Instruction Generation and Augmentation for Zero-Shot Bias Mitigation & F & F & F \\
2024 & Tell Me Why: Explainable Public Health Fact-Checking with Large Language Models & T, E & T, E & E \\
2024 & FairBelief - Assessing Harmful Beliefs in Language Models & F & F & F \\
2024 & The Trade-off between Performance, Efficiency, and Fairness in Adapter Modules for Text Classification & F & F & F \\
2024 & WhenXGBoost OutperformsGPT-4 on Text Classification: A Case Study & --- & --- & --- \\
2024 & Towards HealthyAI: Large Language Models Need Therapists Too & ES & ES & ES \\
2024 & Exploring Causal Mechanisms for Machine Text Detection Methods & E & R, E & R \\
2024 & FactAlign: Fact-Level Hallucination Detection and Classification Through Knowledge Graph Alignment & T & T & T \\
2024 & Cross-Task Defense: Instruction-TuningLLMs for Content Safety & R, ES & R, ES & R, ES \\
2025 & Beyond Text-to-SQLforIoTDefense: A Comprehensive Framework for Querying and ClassifyingIoTThreats & ES & --- & --- \\
2025 & Line of Duty: EvaluatingLLMSelf-Knowledge via Consistency in Feasibility Boundaries & T & T & T \\
2025 & Multi-lingual Multi-turn Automated Red Teaming forLLMs & R & R, ES & R, ES \\
2025 & Rainbow-Teaming for thePolish Language: A Reproducibility Study & R & R, ES & R, ES \\
2025 & BiasEdit: Debiasing Stereotyped Language Models via Model Editing & F & F & F \\
2025 & Do Voters Get the Information They Want? Understanding Authentic VoterFAQs in theUSand How to Improve for Informed Electoral Participation & T & T & --- \\
2025 & ViBe: A Text-to-Video Benchmark for Evaluating Hallucination in Large Multimodal Models & T & T & T \\
2025 & Know What You do Not Know: Verbalized Uncertainty Estimation Robustness on Corrupted Images in Vision-Language Models & T & T & T \\
2025 & Summary the Savior: Harmful Keyword and Query-based Summarization forLLMJailbreak Defense & R & R, ES & R \\
2025 & Bias A-head? Analyzing Bias in Transformer-Based Language Model Attention Heads & F & F, E & F \\
2025 & Mimicking How Humans Interpret Out-of-Context Sentences Through Controlled Toxicity Decoding & F & F & F, ES, E \\
2025 & Gibberish is All You Need for Membership Inference Detection in Contrastive Language-Audio Pretraining & P & P & P \\
2025 & On the Robustness of Agentic Function Calling & R & R & R \\
2025 & MonteCarlo Temperature: a robust sampling strategy forLLM’s uncertainty quantification methods & T & T & T \\
2025 & Know Thyself: Validating Knowledge Awareness ofLLM-based Persona Agents & T & T & T \\
2025 & Building SafeGenAIApplications: An End-to-End Overview of Red Teaming for Large Language Models & R & R, ES & R \\
2025 & Difficulty Estimation in Natural Language Tasks with Action Scores & --- & --- & --- \\
2025 & Are Small Language Models Ready to Compete with Large Language Models for Practical Applications? & --- & --- & --- \\
2025 & A Calibrated Reflection Approach for Enhancing Confidence Estimation inLLMs & T & T & T \\
2025 & Evaluating Design Choices in Verifiable Generation with Open-source Models & T & T & T, E \\
2025 & Battling Misinformation: An Empirical Study on Adversarial Factuality in Open-Source Large Language Models & R, T & R, T & R, T \\
2025 & Will the Prince Get True Love’s Kiss? On the Model Sensitivity to Gender Perturbation over Fairytale Texts & F & F & F \\
2025 & PBI-Attack: Prior-Guided Bimodal Interactive Black-Box Jailbreak Attack for Toxicity Maximization & R & R & R, ES \\
2025 & Disentangling Linguistic Features with Dimension-Wise Analysis of Vector Embeddings & E & E & E \\
2025 & Gender Encoding Patterns in Pretrained Language Model Representations & F & F, E & F \\
2025 & Defining and Quantifying Visual Hallucinations in Vision-Language Models & T & T & T \\
2025 & Revitalizing Saturated Benchmarks: A Weighted Metric Approach for Differentiating Large Language Model Performance & --- & --- & --- \\
2025 & Synthetic Lyrics Detection Across Languages and Genres & --- & --- & --- \\
2025 & A Lightweight Multi Aspect Controlled Text Generation Solution For Large Language Models & ES & --- & F, ES \\
2025 & Gender Bias in Large Language Models across Multiple Languages: A Case Study ofChatGPT & F & F & F \\
2025 & Investigating and Addressing Hallucinations ofLLMs in Tasks Involving Negation & T & T & T \\
2025 & FACTOID:FACtual enTailment fOr hallucInation Detection & T & T & T \\
2025 & Ambiguity Detection and Uncertainty Calibration for Question Answering with Large Language Models & T & T & T \\
2025 & Smaller Large Language Models Can Do Moral Self-Correction & ES & F, ES & F, ES \\
2025 & Error Detection for Multimodal Classification & E & E & T, E \\
2025 & Break the Breakout: ReinventingLMDefense Against Jailbreak Attacks with Self-Refine & R & R, ES & R, ES \\
2025 & Minimal Evidence Group Identification for Claim Verification & T & T & T \\
2025 & Cracking the Code: Enhancing Implicit Hate Speech Detection through Coding Classification & F & F & F \\
2026 & Evaluating Cross-Lingual Behavior and Consistency of Multimodal Large Language Models & F & T & --- \\
2026 & Hair-Trigger Alignment: Black-Box Evaluation Cannot Guarantee Post-Update Alignment & R & --- & R, ES \\
2026 & Teaching PeopleLLM’s Errors and Getting it Right & T & T & ES, T \\
2026 & Linear Probes Detect Task Format, Not Reasoning Mode in Language Model Hidden States & E & E & E \\
2026 & KoLegalQA: AKorean LegalQADataset for Trustworthy and Explanation-Grounded LegalAI & ES, E & T, E & E \\
2026 & Authorization-First Retrieval: Enforcing Least Privilege in Multi-AgentRAGSystems & R & R, P & R, P \\
2026 & PIIJailbreaking inLLMs via Activation Steering Reveals Personal Information Leakage & R, P & R, P & R, P \\
2026 & Coercion Suppression Increases Preference Hallucinations via a Deceptive Bypass inK-Level Negotiation Agents & T & --- & ES, T \\
2026 & Purdah and Patriarchy: Evaluating and MitigatingSouthAsian Biases in Open-Ended MultilingualLLMGenerations & F & F & F \\
2026 & Ghost Context: Measuring Cross-Context Interference in Long-Context Language Models & T & T & T \\
2026 & Through a Compressed Lens: Investigating The Impact of Quantization on Factual Knowledge Recall & T, E & T & T \\
2026 & Understanding the Effects of Safety Unalignment on Reasoning- and Instruction-Tuned Large Language Models & R, ES & R, ES, T & R, ES, T \\
2026 & ReacTOD: Bounded Neuro-Symbolic AgenticNLUfor Zero-Shot Dialogue State Tracking & T & --- & T \\
2026 & Geometric Deviation as an Unsupervised Pre-Generation Reliability Signal: ProbingLLMRepresentations for Answerability & T, E & T, E & T, E \\
2026 & Multilingual Steering by Design: Multilingual Sparse Autoencoders and Principled Layer Selection & E & E & E \\
2026 & Deactivating Refusal Triggers: Understanding and Mitigating Overrefusal in Safety Alignment & R & R, ES & R, ES \\
2026 & A Systematic Taxonomy of Failure Modes in Retrieval-Augmented Generation Systems & --- & T & --- \\
2026 & Improving the Faithfulness ofLLM-based Abstractive Summarization with Span-level Unlikelihood Training & T & T & T \\
2026 & Context MisleadsLLMs: The Role of Context Filtering in Maintaining Safe Alignment ofLLMs & ES & R, ES & R, ES \\
2026 & Lexical Familiarity Predicts Processing Depth for Nonliteral Language in Large Language Models & E & E & E \\
2026 & Did You Forget WhatIAsked? Prospective Memory Failures in Large Language Models & R & --- & --- \\
2026 & Don’t Want YourLLMto Recommend Nuclear Strike? Try Asking It inJapanese & R & ES & ES \\
2026 & Toward Dialect-Aware Safety Evaluation forArabic Large Language Models & F, ES & F, ES & F, ES \\
2026 & Single-Layer Activation Edits Easily Corrupt Factual Recall but Rarely Repair It & T, E & T, E & R, T \\
2026 & Truth or Dare: AnalyzingLLMSusceptibility to External Evidence of Varying Factuality & T & R, T & R, T \\
2026 & Uncertainty-Aware Proxy Attribute Reasoning for Reliable Media Bias Detection & F & F & F \\
2026 & The Halo Effect and Language Takeover: Spatiotemporal Attention Decay Explains Vision-Language Model Failures in Simple Visual Counting & T, E & --- & T, E \\
2026 & Why is "Chicago" Predictive of Deceptive Reviews? UsingLLMs to Discover Language Phenomena from Lexical Cues & E & E & E \\
2026 & Domain-Dependent Safety Behavior in Open-WeightLLMs: An Empirical Study Across Seven Ethical Domains & ES & ES & ES \\
2026 & A Systematic Comparison between Extractive Self-Explanations and Human Rationales in Text Classification & E & T, E & E \\
2026 & Guiding Giants: Lightweight Controllers for Weighted Activation Steering inLLMs & E & ES & ES \\
2026 & SURGELLM: Rethinking Multi-Task Evaluation through Task-Aware Feature Gating with Class-Balanced Normalization & --- & --- & --- \\
2026 & With a Grain ofSALT: AreLLMs Fair Across Social Dimensions? & F & F & F \\
2026 & GateKD: Confidence-Gated Closed-Loop Distillation for Robust Reasoning & T & T & T \\
2026 & QuantifyingLLMSafety Degradation Under Repeated Attacks Using Survival Analysis & R, ES & R, ES & R, ES \\
2026 & The Geometry of Refusal: Linear Instability in Safety-AlignedLLMs & ES, E & R, ES, E & R, ES \\
2026 & The ConservativeAI: Diagnosing Hold Bias and Reliability Limits in Persona-Based Monetary Policy Simulation & ES & --- & F, T \\
2026 & ClaimCLAIRE: A Trust-Aware Multi-Component Fact-Checking Agent for Open-World Claims & T & T & T, E \\
2026 & ChatbotManip: a Dataset to Facilitate Evaluation and Oversight of Manipulative Chatbot Behaviour & ES, T & ES & ES \\
2026 & ControllablePareto Trade-off between Fairness and Accuracy & F & F & F \\
2026 & What are They Thinking? Delineation, Probing, and Tracking of Concepts inLLMs & E & E & E \\
\end{longtable}
}

\end{document}